\documentclass[letterpaper, 10 pt, journal]{IEEEtran/IEEEtran}  

\usepackage{personal}
\usepackage{multicol}
\usepackage{amsmath}
\usepackage{amssymb}
\usepackage{cite}
\usepackage[font=footnotesize]{caption}
\usepackage{algorithmic,algorithm2e,float}

\IEEEoverridecommandlockouts

\newcommand{\wtfargmin}[1]{\underset{#1}{\operatorname{arg}\,\operatorname{min}}\;}
\newcommand{\cupdot}{\mathbin{\mathaccent\cdot\cup}}

\begin{document}
\title{
        Learning Manipulation States and Actions for \\ Efficient Non-prehensile Rearrangement Planning
     }
     \author{Joshua A. Haustein${}^{1, \dagger}$, Isac Arnekvist${}^{1, \dagger}$, Johannes Stork${}^{1}$, Kaiyu Hang${}^{2}$
             and Danica Kragic${}^{1}$
         \thanks{${}^{1}$ Robotics, Perception and Learning Lab (RPL), CAS, EECS,
                    KTH Royal Institute of Technology,
                    Stockholm, Sweden,
                    E-mail: haustein, isacar, jastork, dani@kth.se}
         \thanks{${}^{2}$GRAB Lab, Yale University, New Haven, USA, E-mail: kaiyu.hang@yale.edu}
         \thanks{${}^{\dagger}$ J.A. Haustein and I. Arnekvist contributed equally to this work.}
         \thanks{}{}
       }

    \maketitle
    \begin{abstract}
    This paper addresses non-prehensile rearrangement planning problems where a
    robot is tasked to rearrange objects among obstacles on a planar surface.
    We present an efficient planning algorithm that is designed to impose few
    assumptions on the robot's non-prehensile manipulation abilities and is
    simple to adapt to different robot embodiments. For this, we combine
    sampling-based motion planning with reinforcement learning and generative
    modeling. Our algorithm explores the composite configuration space of
    objects and robot as a search over robot actions, forward simulated in a
    physics model. This search is guided by a generative model that provides
    robot states from which an object can be transported towards a desired
    state, and a learned policy that provides corresponding robot actions. As
    an efficient generative model, we apply Generative Adversarial Networks. We
    implement and evaluate our approach for robots endowed with configuration
    spaces in $SE(2)$. We demonstrate empirically the efficacy of our algorithm
    design choices and observe more than 2x speedup in planning time on various
    test scenarios compared to a state-of-the-art approach.
\end{abstract}

    \section{Introduction}
\label{sec:introduction}
In cluttered environments, a robot frequently encounters situations
in which it needs to rearrange objects.
A planning algorithm needs to reason about
\textit{which} objects to move, \textit{where} to move them and \textit{how} to move them.
To achieve this, one needs to model the robot's manipulation abilities
and how these affect the environment. Many existing approaches simplify this
by limiting manipulation to primitives like pick-and-place or straight-line pushing with
the robot's end-effector. Pushing is commonly further limited to a single
object at a time and it is assumed that the object moves quasistatically, i.e.\ that
inertial forces are neglectable and the object does not slide.

We are interested in enabling robots to utilize a wider range of non-prehensile
manipulation skills. We aim to enable a robot to push
multiple objects simultaneously with any of its parts and not just
its end-effector. Modeling such skills explicitly, however, is labor-intensive
and difficult to adjust to different robot embodiments and objects.
Therefore, in this work, we devise an algorithm for planar non-prehensile rearrangement planning
that does not require manually designed manipulation primitives.
\begin{figure}[t]
       \includegraphics[width=\columnwidth]{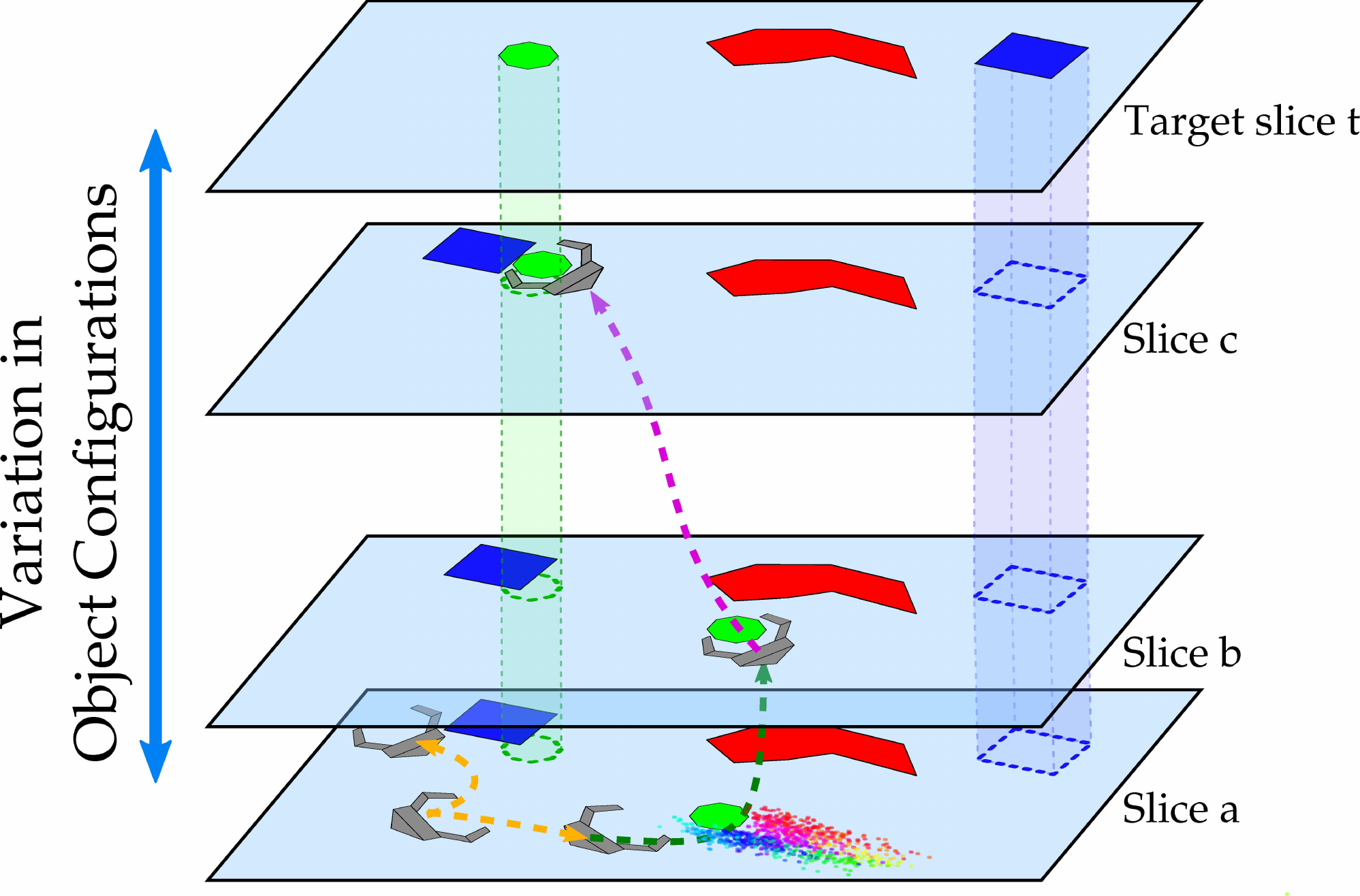}
       \centering
        \caption{
                Our planning algorithm rearranges movable objects on a plane (drawn in green and blue)
                in the presence of obstacles (red) using non-prehensile manipulation.
                The algorithm explores the composite configuration space
                of robot and objects in slices. Each slice corresponds to an object
                arrangement and is the subset of configurations with this arrangement.\\
                \textbf{Slice a:}
                Configurations that lie within the same slice are connected through collision-free
                robot motions (yellow arrows). To purposefully transition towards a target slice $t$,
                our algorithm first selects an object (green) to transport towards the target.
                Thereafter, it samples a robot state from a learned distribution (colored points)
                from which the desired transport is feasible.\\
                \textbf{Slice b:}
                After steering the robot to the sampled state, it applies a learned policy to push
                the object towards the target. The outcome of every robot action is modeled using
                a dynamic physics model, which also allows modeling unintended additional contact.\\
                \textbf{Slice c:}
                The model also models object-object contact,
                which enables the approach to push multiple objects simultaneously.\\
                \textbf{Video:} An illustrative video of this and additional supplementary material
                can be found on \url{https://joshuahaustein.github.io/learning_rearrangement_planning/}
        }
        \label{fig:concept}
\end{figure}

Instead, our approach
combines recent works on physics-based
non-prehensile rearrangement planning~\cite{Cosgun2011, King2015, Haustein2015, King2016} with
reinforcement learning and generative modeling.
Physics-based rearrangement planners explore the composite configuration
space of objects and robot by performing a forward search over simple robot-centric actions.
A physics model is applied to predict whether any action leads to collision
and what the effects of this collision are. This allows the
planner to exploit any manipulation the robot can achieve with its motions.

The key ideas of this work lie in augmenting this physics-based approach
through the following concepts (see also \figref{fig:concept}):
\begin{itemize}
    \item[1.] We provide the algorithm with learned guidance on how to manipulate objects.
        For this, we train a generative model to efficiently produce samples from the set of robot configurations
        from which transporting an object towards a desired state is feasible. We achieve this
        by applying Generative Adversarial Networks (GANs). Additionally, we
        train a one-step policy that, once the robot is placed in such a
        configuration, provides object-specific pushing actions.
    \item[2.] We structure the search by segmenting the search space into sets of configurations with
        similar object arrangements. This acknowledges that steering the robot is simpler
        than rearranging objects, and allows the algorithm to efficiently select suitable
        nearest neighbors for its search tree extension.
\end{itemize}
As a physics-based rearrangement planner, our approach can compute solutions
where the robot pushes multiple objects simultaneously, as often as needed, and with any
of its parts. In contrast to related approaches, our
algorithm is more efficient and easily adapted to different object types
and robot geometries.

We implement our approach for robots endowed with configuration spaces in $SE(2)$,
and evaluate it experimentally.
We observe the efficacy of our design choices and achieve more than 2x planning
speedup compared to a state-of-the-art physics-based approach on various test scenarios.

The remainder of this work is structured as follows. First, we formalize the
problem we address in \secref{sec:problem} and summarize related works in
\secref{sec:related_works}. Then, in \secref{sec:concept} we provide a more detailed
overview of our concepts and contributions.
We present the details of our approach in \secref{sec:method}, and evaluate
it in \secref{sec:experiments}. We discuss limitations and future
work in \secref{sec:discussion}.

    \section{Problem Definition \texorpdfstring{$\And$}{And} Notation}
\label{sec:problem}
We consider a rearrangement planning problem where a robot
is tasked to rearrange multiple objects using only non-prehensile manipulation.
The robot operates in a bounded environment that contains $m$ movable objects among
static obstacles $\mathcal{O}$. Its goal is to rearrange a target set~$\mathbf{T} \subseteq \{1, \ldots, m\}$
of movable objects to some goal configurations~$\{g_i~|~i \in \mathbf{T}\}$.
This target set may either encompass all $m$ objects or only a subset.
To achieve its goal, the robot may manipulate any movable object while avoiding
collisions with the static obstacles. We assume that all objects are rigid bodies that are placed on a
single planar support surface and we disallow toppling over objects.

We formulate this problem as a motion planning problem on the composite configuration space
of all movable objects and the robot, $\mathcal{C}_{0:m} = \mathcal{C}_0 \times \cdots \times \mathcal{C}_m$.
Here, $\mathcal{C}_0$ is the configuration space of the robot and
$\mathcal{C}_i \subset \mathit{SE}(2)$ the configuration space of movable object $i \in \{1, \ldots, m\}$.
The composite configuration space is partitioned in free and obstacle space,
$\mathcal{C}_{0:m}~=~\mathcal{C}_{0:m}^\text{free}~\cupdot~\mathcal{C}_{0:m}^\text{obst}$. We define
the free space $\mathcal{C}_{0:m}^\text{free}$ as the set of physically feasible configurations
in which no two objects or the robot overlap and there is no collision with any static obstacle.
Note that this definition includes configurations in which there is contact among
movable objects or between these and the robot.
For a composite configuration $x = (x_0, \ldots, x_m)~\in~\mathcal{C}_{0:m}$ we refer by
$x_i \in \mathcal{C}_i$ to the configuration of object~$i \geq 1$ and by $x_0 \in \mathcal{C}_0$ to the
configuration of the robot. We use the terms \textit{state} and \textit{configuration} interchangeably and
distinguish between two different states using the superscript notation $x^a$, $x^b$.
Our notation is summarized in \tableref{table:notation}.

\begin{table}
    \resizebox{\columnwidth}{!}{%
        \begin{tabular}{l|p{0.45\columnwidth}}
    \textbf{Symbol} & \textbf{Description} \\ \hline
        $m$ & Total number of movable objects\\
        $\mathcal{O}$ & Static obstacles\\
        $\mathcal{C}_0$ & Configuration space of the robot \\
        $\mathcal{C}_i$ & Configuration space of object $i$ \\
        $\mathcal{C}_{0:m}$ & Composite configuration space \\
        $\mathcal{C}_{1:m}$ & Space of all object arrangements \\
        $\mathcal{C}_{0:m}^\text{free}$ & Feasible configuration space\\
        $\mathcal{C}_{0:m}^\text{obst}$ & Obstacle space\\
        $x = (x_0, \ldots, x_m)$ & Composite state \\
        $x_{1:m} = (x_1, \ldots, x_m)$ & Composite state of objects \\
        $x^a, x^b$ & Two different composite states \\
        $x_i$ & State of object $i$ \\
        $x_0$ & State of the robot \\
        $x_i^a, x_i^b$ & Two different states of object $i$ \\
        $u \in U$ & Robot action in action space $U$ \\
        $\mathbf{T} \subseteq \{1, \ldots, m\}$ & Target object set \\
        $\mathcal{G} \subset \mathcal{C}_{0:m}^\text{free}$ & Goal region \\
        $g_i \in \mathcal{C}_i$ & Goal configuration of object $i$ \\
        $\xi: [0, t_\text{max}] \to \mathcal{C}_{0:m}^\text{free}$ & Path in $\mathcal{C}^\text{free}_{0:m}$ \\
        $\tau: [0, t_\text{max}] \to U$ & Time-action mapping \\
        $d_{\mathcal{C}_i}:\mathcal{C}_i \times \mathcal{C}_i \to \mathbb{R}$  & Distance function on $\mathcal{C}_i$ \\
        $\Gamma: \mathcal{C}_{0:m} \times U \to \mathcal{C}_{0:m}$ & Deterministic physics model \\
        $\bot$ & Empty value or invalid state\\
        $s$ & A subset of $\mathcal{C}_{0:m}$ (a slice) with the same object arrangement\\
        $s_{1:m}$ & Object arrangement in slice $s$\\
        $s_{i}$ & Object i's state in slice $s$\\
        $d_{s}:\mathcal{C}_{1:m} \times \mathcal{C}_{1:m} \to \mathbb{R}$  & Distance function on  $\mathcal{C}_{1:m}$ \\
        $G_\varphi$ & Robot state generator \\
        $\pi_\theta$ & Pushing policy\\
        $L_\theta$ & Policy loss function\\
        $\mathcal{T}$ & Search tree of states in $\mathcal{C}_{0:m}^\text{free}$\\
        $\mathcal{S}$ & Set of explored slices of $\mathcal{C}_{0:m}^\text{free}$\\
        $\nu_i$ & Physical properties of object $i$ \\
        $x_i^d$ & Desired state for object $i$ \\
        $x_i'$ & Resulting state for object $i$ \\
    \end{tabular}}
    \caption{Notation used in this work.}
    \label{table:notation}
\end{table}

The robot can interact with its environment through actions $u \in U$.
The purpose of our algorithm is to compute a time-action mapping $\tau:~[0,~t_\text{max}]~\to~U$
and a path $\xi:~[0, t_\text{max}]~\to~\mathcal{C}_{0:m}^\text{free}$,
where $t_\text{max} \geq 0$ is the duration of this solution.
The path $\xi$ describes which states the system transitions through
and the time-action mapping $\tau$ which action to execute at time
$t \in[0, t_\text{max}]$. Such a pair $(\xi, \tau)$ is physically feasible,
if the non-holonomic constraint $\dot{\xi}(t)~=~f_\text{physics}(\xi(t),~\tau(t))$ is
fulfilled for all $t \in [0, t_\text{max}]$. This constraint
expresses the fact that objects do not move by themselves,
but only as a result of forces exerted on them by the robot.

Given an initial state of the environment $x^0~\in~\mathcal{C}_{0:m}^\text{free}$
and a goal region~$\mathcal{G} \subseteq \mathcal{C}_{0:m}^\text{free}$,
the goal of our algorithm is to find a physically feasible tuple $(\xi, \tau)$ such
that $\xi(0) = x^0$ and $\xi(t_\text{max}) \in \mathcal{G}$.
We define $\mathcal{G} \subseteq \mathcal{C}_{0:m}^\text{free}$
as the set of states for which all target objects
$\mathbf{T}$ are located close to goal configurations $g_i \in \mathcal{C}_i$, i.e.
$$\mathcal{G}~=~\{x~\in~\mathcal{C}_{0:m}^\text{free}~|~\forall~i~\in \mathbf{T} : d_{\mathcal{C}_i}(x_i, g_i) \leq \epsilon_i \}$$
for some thresholds $\epsilon_i > 0$ and distance functions
$d_{\mathcal{C}_i}:~\mathcal{C}_i~\times~\mathcal{C}_i~\to~\mathbb{R}$.

    \section{Related Works}
\label{sec:related_works}
Our problem definition falls into the category of manipulation planning.
Manipulation planning involves both planning the motion of a robot
and the manipulation of one or multiple objects in the presence of obstacles.
Conceptually, we can distinguish between related works that focus
on manipulating a single object
from approaches that manipulate
multiple objects. Manipulation planning for multiple objects can be further subdivided into
manipulation planning among movable obstacles (MAMO)~\cite{Stilman2007, Dogar2012},
rearrangement planning (RP)~\cite{Wilfong1991, Alami1994, Ben-Shahar1998b, Ota2004, Scholz2010, Krontiris2015, Krontiris2016, Garrett2018, Barry2013a, Havur2014, Jentzsch2015}
and navigation among movable obstacles (NAMO) \cite{Stilman2005, VandenBerg2009,Nieuwenhuisen2008}.

These subcategories differ in the definition
of the goal of the task. Applying our terminology, MAMO defines the target object set $\mathbf{T}$, or
alternatively $\mathcal{G}$, to only encompass a single object. The remaining
movable obstacles may need to be rearranged to succeed at the task, but their final
configurations are not relevant.
In contrast, in RP the target set $\mathbf{T}$ encompasses all movable objects, i.e.\ the task
is to rearrange all objects to specific locations. Finally, in NAMO the goal is to navigate the
robot to a goal configuration when clearing the path from obstacles
is required. In this case, the goal is only defined for the robot, i.e.\ $\mathbf{T} = \{0\}$, and
the final configurations of the movable obstacles are not of interest.

Since we allow $\mathbf{T}$ to be any subset of $\{1, \ldots, m\}$, our problem covers
manipulation planning for a single object, MAMO and RP, but not NAMO.
Our work addresses the special cases of these problem classes
when all objects are located on a single support surface
and the robot only applies non-prehensile manipulation. For brevity, we refer
to our problem simply as a rearrangement planning problem.

\subsection{Complexity and Concepts}
\label{sec:related_works:concepts}
Manipulation planning and in particular rearrangement planning is challenging.
Wilfong et al. \cite{Wilfong1991} showed that NAMO is NP-hard, and that
rearrangement planning is PSPACE-hard. The complexity arises from the high dimensional
search space and the constraint that objects only move as a consequence
of the robot's actions. Alami et al.~\cite{Alami1994} therefore classify robot
actions into two categories. \textit{Transit} actions are collision-free robot motions, whereas
\textit{transfer} actions denote actions that manipulate objects. Planning
\textit{transit} actions is a classical robot motion planning problem, whereas
planning \textit{transfer} actions requires additional reasoning about the mechanics of manipulation.

Depending on the type of action, a manipulation planner operates on different lower dimensional
subspaces of the configuration space $\mathcal{C}_{0:m}$~\cite{Alami1994, Ben-Shahar1998a, Simeon2004}.
If a robot moves without contact, it transitions between states within a subspace of
$\mathcal{C}_{0:m}^\text{free}$ for which all objects are at the
same location. When, for instance, the robot grasps an object, it transitions to a different subspace,
in which the grasped object and the robot move simultaneously.
Other types of manipulation implicitly define different subspaces with different transition points.
The challenges of manipulation planning lie within computing motions within
each of these subspaces, finding transition points between them, as well as
computing a global path that moves via multiple subspaces towards a goal.

\subsection{Manipulation Planning with Pick-And-Place}
Many works on manipulation planning apply pick-and-place
for \textit{transfer} actions \cite{Alami1994, Simeon2004, Ota2004, Stilman2007, Krontiris2014,
Krontiris2015, Krontiris2016, Garrett2018}. From a planning perspective this is attractive for two reasons.
First, it restricts where a transition from a \textit{transit}
to a \textit{transfer} action can occur, i.e.\ at the configurations where the robot
grasps a resting object.
Second, it simplifies modeling how a state $x = (x_0, \ldots, x_m)$ evolves during a
\textit{transfer} action. Assuming the robot grasps object $k$, a pick-and-place operation
moves $x$ to $x'$ such that only $x_0 \neq x_0'$ and $x_k \neq x_k'$, while
all other objects remain at the same location, i.e.\ $x_i' = x_i$. Furthermore, the new object
state $x_k'$ can be directly computed from the new robot state $x_0'$
given a known transformation between the grasped object and the robot's end-effector's frame.

The early work by Alami et al. \cite{Alami1994} presents two algorithms building on this.
The first algorithm addresses rearrangement planning of multiple objects and assumes a finite set of grasps and
placements. The second addresses manipulation planning for a single
polygonal object under continuous grasps. Simeon et al. \cite{Simeon2004} adapted this approach to
manipulation planning for a single object under continuous grasps and
placements using probabilistic roadmaps.

Later, Stilman et al.~\cite{Stilman2007} addressed the problem of MAMO using pick-and-place.
The algorithm first computes which objects collide with the desired pick-and-place
operation on the target object and then recursively attempts to clear these obstructions.
To limit the recursion depth, the approach assumes \textit{monotone} problems,
where each obstructing object needs to be moved at most once.

Krontiris et al. \cite{Krontiris2015} integrate a modification of this MAMO algorithm
as local planner into a bidirectional RRT algorithm~\cite{lavalle06book} to solve rearrangement problems.
The RRT algorithm explores the space of object arrangements and utilizes the MAMO
algorithm to compute pick-and-place sequences to transition between different arrangements.
More recently~\cite{Krontiris2016}, the same authors augmented this approach
using minimal constraint removal paths~\cite{Hauser2014} to compute an order in which to move
the objects that minimizes obstructions, thus avoiding the extensive backtracking of the original MAMO algorithm.
This results in a non-monotone rearrangement planning algorithm that
was demonstrated to solve even challenging problems with many objects efficiently.

\subsection{Manipulation Planning with Diverse Manipulation Types}
\label{sec:related_works:diverse}
A robot can rearrange objects more effectively and efficiently when it is not only limited to
pick-and-place, but can also apply non-prehensile manipulation like pushing.
Pushing is useful as it is faster to execute,
can reduce uncertainty \cite{Dogar2010} and can be applied to heavy or multiple objects simultaneously.
Hence, several existing works aim at combining diverse types of
manipulation \cite{Dogar2012, Barry2013, Barry2013a, Havur2014, Jentzsch2015} in a single planner.

In these works, the different types of manipulation are abstracted as
manipulation primitives. A manipulation primitive defines what its preconditions
and effects are. For instance, for a picking primitive the precondition
is that the robot's end effector is located at a grasp pose relative to a target object.
Its effect is then, as described above, the change of a single object's pose
and the robot's configuration.
Similarly, one can formalize primitives for pushing. Here, however, in
general both the preconditions and effects are more challenging to model.
A robot may push an object $k$ with its various parts from many different
robot configurations.
Furthermore, the pushed object may slide or collide with other objects, resulting
in state changes for these as well. This renders modeling
the resulting state $x'$ of a pushing operation challenging.

To simplify this, the works in this category commonly limit the implemented pushing primitives
to cases where a single object is pushed with the robot's end-effector. The effect
of pushing is often assumed to be simply the translation of the robot's end-effector
applied to the pushed object.

More sophisticated pushing primitives are presented by Dogar et al.~\cite{Dogar2012}.
The authors apply a quasistatic
pushing model~\cite{Lynch1996} to compute capture regions of different
pushing primitives. These capture regions describe sets of object poses relative to a
robot link, e.g.\ the palm of a robotic hand. If an object is located within
the capture region of a primitive, it is guaranteed to end up
inside a known subregion of this region once the primitive was executed.
This allows to compute solutions for MAMO under object pose uncertainty,
but is limited by the quasistatic assumption that inertial forces are negligible.
Furthermore, the approach is limited to manipulating a single object at a time.

\subsection{Manipulation Planning with Pushing}
There have also been works that solely focus on manipulation planning for
pushing~\cite{Lynch1996, Nieuwenhuisen2005, Zhou2017}.
Lynch et al.\ \cite{Lynch1996} studied the controllability
of a single object pushed with point or line contacts
under the quasistatic assumption.
The authors present a planner based on Dijkstra's
algorithm to plan pushing a single object among static obstacles
using stable line contacts that can be executed open-loop.
More recently, Zhou et al.\ \cite{Zhou2017} showed that quasistatic push planning of an object
using a single contact can be reduced to computing a Dubins path. The authors applied
this in the RRT algorithm to plan pushing paths that avoid obstacles with the
pusher and the object.  While these approaches produce very accurate solutions
that can be executed open-loop, they are limited to situations where the
quasistatic assumption holds and only a single object is pushed.

In rearrangement planning it may be more efficient to push multiple
objects simultaneously than in sequence. Ben-Shahar et al.\
\cite{Ben-Shahar1998a, Ben-Shahar1998b} present a rearrangement planner that, in principle,
allows pushing objects simultaneously.  The approach performs a hill-climbing
search on a grid-discretization of the composite configuration space. The
search minimizes a cost function that denotes the minimal cost to reach the
goal. This cost is precomputed on the discrete search space using a reverse
pushing model that could take object-object contacts and non-quasistatic
physics into account. Obtaining such a general reverse pushing model, however, is
difficult and the authors present only a simplified model in their experiments.

\subsubsection{Integrating a Physics Model}
Over the last decades rigid-body physics simulators have become widely
available~\cite{bullet, box2d, ode, mujoco}. These simulators
can model the physical interactions between a robot and its surrounding objects.
Integrating such a physics model with a motion planning algorithm
allows to predict the effects of robot actions.

Zito et al.\ \cite{Zito2012} use this approach to plan to push a single object.
Kitaev et al.\ \cite{Kitaev2015} combine a physics model with stochastic
trajectory optimization to compute grasp approach motions in clutter. A similar
idea has been applied to rearrangement planning and MAMO in \cite{King2015} and \cite{Haustein2015}.
These works integrate a physics model
into the kinodynamic RRT algorithm~\cite{LaValle2001}.
In \cite{King2015}, a quasistatic multi-body pushing model is used to
generate solutions where the robot utilizes its full-arm to push multiple object simultaneously.
In \cite{Haustein2015}, instead a dynamic physics model is used that allows
generating solutions where a robot pushes and thrusts sliding objects into place.

In both \cite{King2015} and \cite{Haustein2015} the RRT algorithm explores
$\mathcal{C}_{0:m}$ by forward propagating randomly selected robot actions through the
physics model. These robot-centric actions move the robot
and make no explicit assumption on how an object may be manipulated.
As a consequence, both approaches can achieve a diverse set of manipulation actions, but
suffer from a poor exploration of the search space.
Consider two states $x^a, x^b~\in~\mathcal{C}_{0:m}$ for which some object state differs,
i.e. $x^a_k \neq x^b_k$ for an object $k$.
Without any object-centric pushing primitives, the algorithms lack guidance on how to
select an action that takes the search from $x^a$ closer to $x^b$.
Randomly sampling robot-centric actions may lead to incidental contact with an object,
but it is very unlikely to sample an action that moves the object towards the desired state.
This diminishes the RRT's property of rapidly exploring the search space.

King et al.~\cite{King2016} address this by augmenting the approach from \cite{King2015}
with object-centric pushing primitives. These primitives
are similar to the ones described in \secref{sec:related_works:diverse}, but their effects are modeled
using the physics model. This in combination with random robot-centric actions
significantly improves the algorithm's performance.
However, designing these primitives manually still poses a limiting factor.
First, providing the algorithm with diverse primitives such as pushing with the elbow, or pushing
with the forearm is labor-intensive. Second, adapting the planner to a different robot generally
requires the human designer to create new primitives. Third, selecting the most
suitable primitive within the planner efficiently becomes challenging for a large set of
primitives.

\subsubsection{Learning for Pushing}
\label{sec:related_works:learning}
An alternative to modeling pushing analytically is to apply machine learning.
The works by Scholz et al.~\cite{Scholz2010} and Elliot et al.~\cite{Elliott2016}
learn probabilistic forward models for manually designed manipulation primitives to rearrange objects.
As these forward models are learned from the real world, the approaches make few assumptions
on the outcome of the physical interactions. Object-to-object contacts, however,
are not modeled. Furthermore, the use of manipulation primitives limits the
type of contact the robot applies.

Another probabilistic model-based approach was proposed by Wang et
al.~\cite{Wang2017} and addresses push planning for a single object
under uncertainty. The authors propose to first learn a distribution of
stochastic pushing outcomes in a world containing no obstacles. Then, given
the task to push the object to some goal in the presence of obstacles, the
approach first constructs several search trees of collision-free object poses
using the RRT algorithm. The nodes of these search trees are then treated as the states of a
finite state Markov decision process (MDP), where the transition probabilities
are given by the pushing distribution learned beforehand. In this MDP, an
optimal value function and policy are then computed iteratively to solve the
given problem. This approach has the benefit that the learned pushing policy
is robust against uncertainty. Also, being model-based reduces the amount of
data needed to train the policy \cite{atkeson1997comparison}. The approach,
however, does not address how to plan the motion of the robot, and instead
assumes that the robot can always transition to any pose relative to the
object. Besides, the value function and policy are based on a dynamics model
which is only valid for the single trained object. Pushing a different object
requires learning a new model.

To remove restrictions on the type of contact, and the type of objects, Finn et
al. \cite{finn2017deep} present an end-to-end approach for pushing. Instead of
explicitly modeling the manipulated object, the approach learns probabilities
of how pixels move in a camera image as a result of robot actions. These learned
probability distributions are, however, only informative for short time scales,
and are unlikely to generalize to longer time-scales and planning in complex
scenes.

In contrast to these model-based approaches, Andrychowicz et
al.~\cite{andrychowicz2017hindsight} apply Q-learning to learn a policy for
pushing a single object in an obstacle-free environment. As the previously
mentioned works \cite{finn2017deep} and \cite{Wang2017}, this makes
few to no assumptions on the type of contact.  Being model-free also means
that the dependence on a correct approximation of the dynamics is no longer
needed. Any changes to the environment or the pushed object, however, requires
training the policy from scratch again.

\subsection{Our Approach}
The goals of this work are to devise an algorithm that
\begin{itemize}
    \item computes solutions to the problem defined in \secref{sec:problem},
    \item makes few limiting assumptions on how the robot can manipulate objects,
    \item maintains planning efficiency.
\end{itemize}
The ideas from \cite{King2015} and \cite{Haustein2015} address the first two
goals by integrating a physics model into the kinodynamic RRT algorithm.
Both approaches, however, suffer from a poor exploration rate, which leads to
long planning times. King et al. \cite{King2016} demonstrated how the efficiency of this family of
algorithms can be improved by equipping them with object-centric actions. The proposed action
primitives, however, require a human designer and thus reintroduce limiting assumptions
on the robot's manipulation abilities. The learning-based approaches presented in
\secref{sec:related_works:learning} either also require some primitives or
do not allow manipulating multiple objects at a time.

    \section{Concept And Contributions}
\label{sec:concept}

As stated in \secref{sec:introduction}, we adopt the ideas from
\cite{King2016, King2015, Haustein2015}. Similar
to these works, we present a sampling-based planning algorithm that explores
the composite configuration space $\mathcal{C}_{0:m}$ as a forward search over simple
robot-centric actions $u \in U$.
To predict the outcome of these actions, we apply a dynamic physics model $\Gamma$.
This, on the one hand, guarantees that the algorithm computes physically feasible solutions.
On the other hand, it shifts the assumptions on the effects of the actions from
the algorithm to the model. For instance, using a dynamic physics simulator as a
model allows our approach to compute solutions where the robot manipulates rolling or sliding
objects, or exploits object-to-object contacts.

In order to avoid
planning in a state space that includes velocities, however, we follow the
approach from \cite{Haustein2015} and limit our search to sequences of dynamic
actions between statically stable states. This means that all objects and the robot are
required to come to rest after each action within some time limit.
Formally, we assume the physics model $\Gamma$ to be a function
$\Gamma: \mathcal{C}_{0:m}^\text{free} \times U \to \mathcal{C}_{0:m}^\text{free} \cup \bot$ that
maps a configuration $x \in \mathcal{C}_{0:m}^\text{free}$ and an action $u$
to the resulting state $\Gamma(x, u) = x' \in \mathcal{C}_{0:m}^\text{free}$ of this action, or a special
value $\bot$, if the action leads to an invalid, i.e.\ non-resting, state.

We augment this physics-based search through two key concepts.
First, we provide the algorithm with guidance on how
to achieve \textit{transfer} actions using machine learning.
This enables our algorithm to compute object-centric actions without manually
designed primitives.
Second, we segment the search space $\mathcal{C}_{0:m}$, and accordingly the search tree,
in sets with similar object arrangements. This structures the search
and results in a more efficient exploration of $\mathcal{C}_{0:m}$ than
by the RRT algorithms in~\cite{King2015, Haustein2015, King2016}.

\begin{figure}[t]
       \includegraphics[width=\columnwidth]{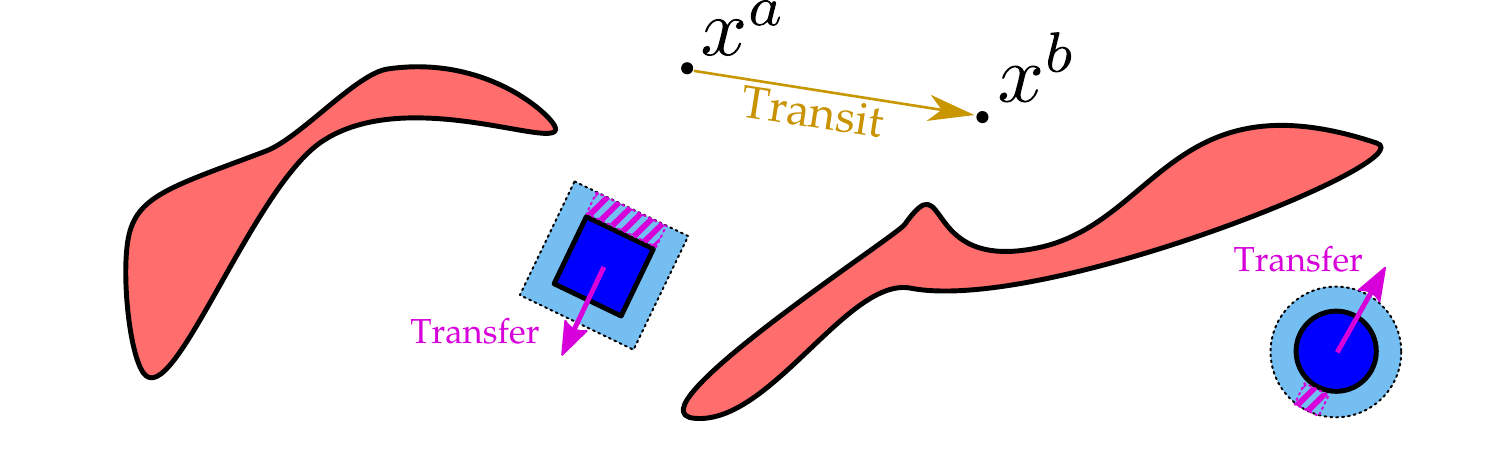}
       \centering
        \caption{
                An illustration of a slice of the configuration space $\mathcal{C}_{0:2}$
                for a fixed object arrangement for a point robot. The red
                areas show parts of $\mathcal{C}_{0:2}^\text{obst}$ that are inaccessible due to collisions
                with static obstacles. The dark blue areas show parts that are in collision with
                movable objects. The yellow arrow indicates a collision-free
                \textit{transit} movement between two states $x^a, x^b$.
                The light blue areas show subsets of $\mathcal{C}_{0:2}$
                from which pushing the objects, i.e.\ a \textit{transfer} action,
                is possible given a particular action space $U$.
                If we intend to push an object in some desired direction (pink arrows), these
                subsets are further restricted (pink striped areas).
               }
        \label{fig:slice_configuration_space}
\end{figure}
To illustrate these concepts, consider \figref{fig:slice_configuration_space}.
It shows an illustration of a cross section of the configuration space $\mathcal{C}_{0:2}$ for a
holonomic point robot and $m=2$ objects. We refer to such a cross section of states
$\{ x~\in~\mathcal{C}_{0:m} | x_{1:m}~=~x^f_{1:m}\}$ with a fixed object arrangement
$x^f_{1:m} = (x^f_1, \ldots, x^f_m)$ as slice of $\mathcal{C}_{0:m}$.
Navigating a robot within a slice, i.e.\ performing \textit{transit} actions,
is a classical motion planning problem
that is well studied~\cite{lavalle06book}. In particular,
steering the robot between two states
$x^a = (x_0^a, x^f_1, \ldots, x^f_m)$, $x^b = (x_0^b, x^f_1, \ldots, x^f_m)$
within the same slice ignoring obstacles is simple for many types of robots.
Navigating to a different slice, i.e.\ performing \textit{transfer} actions,
however, requires the robot to manipulate objects.

\begin{figure*}[!ht]
       \includegraphics[width=\textwidth]{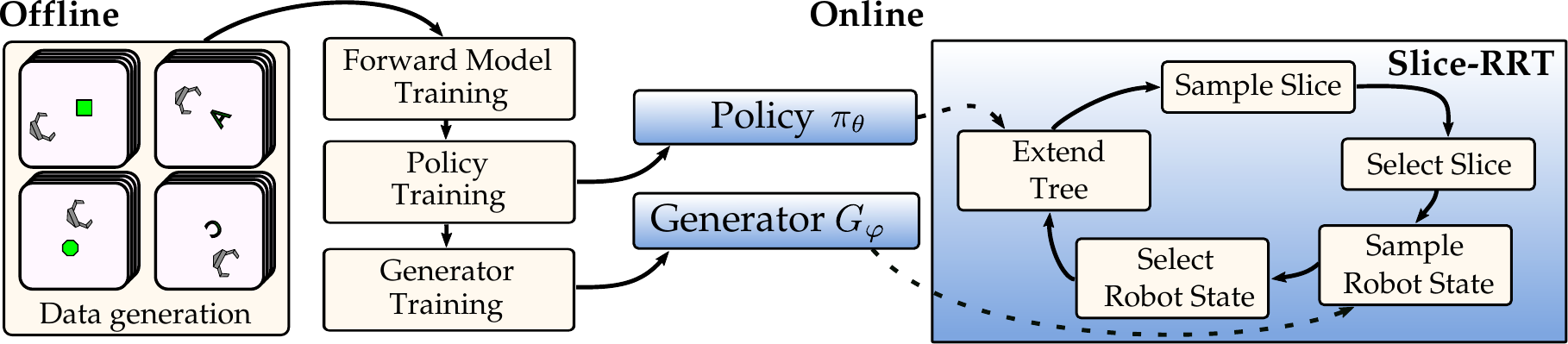}
       \centering
       \caption{
           Our approach consists of an offline and online phase. In the offline phase,
           a policy $\pi_\theta$ and a state generator $G_{\varphi}$ are learned from data generated
           with a physics model. The generator provides samples of robot states that allow transporting
           an object towards a desired state with a single action.
           The policy provides such robot actions. Online, we apply a modified version
           of the RRT algorithm to search for a rearrangement solution. The algorithm explores
           the search space in slices of similar object arrangements. To purposefully transition
           between different slices, it applies the policy and generator to push objects towards
           target states.
        }
        \label{fig:system_overview}
\end{figure*}

The set of robot configurations from which this is achievable
within one action $u \in U$ is shown in \figref{fig:slice_configuration_space} in light blue.
This set is further restricted, if we intend
to push an object in a particular direction~(pink regions).
We refer to these states as \textit{manipulation states}.

The shape of the set of manipulation states depends on the shape of the object, the
action space $U$, and the robot's geometry.
Rather than approximating this manually through
primitives, we train a generative model to produce samples of manipulation states.
In order to transport an object to a desired state, our algorithm then first steers the robot
to such a sampled manipulation state. Then, once the robot is placed in a manipulation state,
the algorithm selects an action $u \in U$ to transport the object towards the desired state.

Selecting this action is generally also non-trivial. The most suitable
action depends on physical properties of the object and the state of the robot.
Hence, we additionally train a policy to provide this action to the planner.
With these two components in place,
our algorithm proceeds as illustrated in \figref{fig:concept}
and explores $\mathcal{C}_{0:m}$ slice-by-slice.

To summarize, the contributions of this work are:
\begin{itemize}
    \item[1.] An efficient sampling-based algorithm for rearrangement planning that is
        agnostic to the details of manipulation.
    \item[2.] An approach to learn manipulation states and actions
    using reinforcement learning and Generative Adversarial Networks (GANs).
    \item[3.]An implementation and evaluation of this framework for robots endowed with
        configuration spaces in $SE(2)$.
\end{itemize}

    \section{Method}
\label{sec:method}

\begin{algorithm}[ht]
    \algsetup{linenosize=\tiny}
    \small
    \SetFuncSty{textsc}
    \DontPrintSemicolon
    \SetKw{Not}{not}
    \SetKw{Is}{is}
    \SetKwInOut{Constants}{Constants}
        \KwIn{Start configuration $x^0 \in \mathcal{C}_{0:m}^\text{free}$,
              maximum number of iterations $n_{\text{max}}$,
              goal region $\mathcal{G} \subset \mathcal{C}_{0:m}^\text{free}$}
    \KwOut{Solution $[(x^0, u^0) \ldots, (x^n, \bot)]$ or $\emptyset$}
    \SetKwFunction{Tree}{Tree}
    \SetKwFunction{Slice}{Slice}
    \SetKwFunction{Add}{AddToSlice}
    \SetKwFunction{SampleSlice}{SampleSlice}
    \SetKwFunction{SelectNode}{SelectNode}
    \SetKwFunction{ExtendTree}{ExtendTree}
    \SetKwFunction{ExtractPath}{ExtractAndShortcut}
    \SetKwFunction{SampleUniform}{SampleUniform}
    \SetKwFunction{SamplePushingState}{QueryGenerator}
    \SetKwFunction{SampleBiased}{RandomChoice}

        $s^0 \gets $ \Slice{$x^0$} \tcp*[f]{Obtain the slice that $x^0$ lies in} \;
        $\mathcal{S} \gets \{s^0\}$ \tcp*[f]{Set of all explored slices} \;
        $\mathcal{T} \gets$ \Tree{$x^0$} \tcp*[f]{Tree of explored states} \;
    \For{$j \gets 1, \ldots, n_{\text{max}}$} {
        $s^d \gets$ \SampleSlice{} \;
        $s^j \gets \wtfargmin{s \in \mathcal{S}} d_s(s, s^d)$ \tcp*[f]{Closest explored slice} \label{algo:line:closest_slice} \;
        $t \gets $ \SampleBiased{$\{0\} \cup \{1, \ldots, m\}$} \;
        \If{$t = 0$} {
            $x_0^d \gets $ \SampleUniform{$\mathcal{C}_{0}^\text{free}$} \;
        }
        \Else {
            $x_0^d \gets $ \SamplePushingState{$s^j, s^d, t$} \;
        }
        $x^j \gets \wtfargmin{x \in s^j} d_{\mathcal{C}_0}(x_0, x_0^d)$ \tcp*[f]{Closest state in $s^j$} \label{algo:line:closest_state}\;
        $x^d \gets (x_0^d, s^d_1, \ldots, s^d_m)$ \tcp*[f]{Constructed target state} \;
        $\mathcal{T}, \mathcal{S} \gets$ \ExtendTree{$x^j, x^d, t, \mathcal{T}, \mathcal{S}$} \;
        \If{$\mathcal{T} \cap \mathcal{G} \neq \emptyset$} {
        \Return \ExtractPath{$\mathcal{T}$} \;
        }
    }
    \Return $\emptyset$ \;
        \caption{The rearrangement planning algorithm.}
    \label{algo:planner_main}
\end{algorithm}
An overview of our approach is shown in \figref{fig:system_overview}.
It consists of three key components: the planning algorithm, the generative model $G_{\varphi}$
and the pushing policy $\pi_\theta$. The planner computes a solution for an instance of our
problem when queried online.
Both the policy and the generator are trained offline and parameterized so that
they can be applied to objects of different shape, mass and friction
properties.

\subsection{Planning Algorithm}
\label{sec:method:planner}
The planning algorithm is outlined in \algref{algo:planner_main}.
We first provide a high-level overview of this algorithm
and then describe the details of its subroutines in the following subsections.

\subsubsection{Overview}
Given a start state $x^0 \in \mathcal{C}^\text{free}_{0:m}$,
the algorithm starts with initializing its search tree $\mathcal{T}$ and a set
$\mathcal{S}$, which keeps track of the slices of $\mathcal{C}_{0:m}^\text{free}$
that the explored states in $\mathcal{T}$ lie in. It then iteratively grows the search tree
$\mathcal{T}$ on $\mathcal{C}^\text{free}_{0:m}$ in a similar fashion to the
kinodynamic RRT algorithm until it either adds a goal state $g \in \mathcal{G}$
to the tree or a maximum number of iterations $n_\text{max}$ is reached.

In each iteration the algorithm needs to select where to extend its search
tree from. For this, it first randomly samples a slice $s^d$ by
sampling a random object arrangement. Next, it selects
the explored slice $s^j \in \mathcal{S}$ that is closest to $s^d$ w.r.t.\ some distance function $d_s$.
For the tree extension the algorithm will either attempt to perform a
\textit{transit} action within the selected slice $s^j$ or a \textit{transfer}
action towards the sampled slice $s^d$.
To decide this, it randomly chooses an index $t \in \{0, \ldots, m\}$.
In case it chooses the robot, $t=0$, the algorithm will attempt a
\textit{transit} action and uniformly samples a robot state $x_0^d$ as a goal for this transit.
In case it chooses an object, $t~\in~\{1, \ldots, m\}$,
it samples a robot state $x_0^d$ that likely allows it to push object $t$ from its
state $s_t^j$ in the current selected slice to its state $s_t^d$ in the randomly sampled slice.
For this, it applies the generator $G_{\varphi}$ that is described in more detail later.

In both cases, the algorithm then needs to select an actual composite state $x^j \in \mathcal{T}$
from the search tree $\mathcal{T}$ to extend from. Since it already selected the slice $s^j$ for extension,
it only needs to search for this state within this slice.
For this, it searches for the state $x^j \in \mathcal{T} \cap s^j$ that minimizes the robot state distance
$d_{\mathcal{C}_0}$ to the sampled robot state $x_0^d$.
Depending on the choice of $t$, the extend function \textsc{ExtendTree} will then either simply steer the robot
to $x_0^d$, or additionally attempt to push object $t$ towards $x_t^d$ using the policy $\pi_\theta$.

The above proceeds until the loop is terminated. If a goal state $g \in \mathcal{G}$ was
added to the tree, the algorithm returns a solution $(\xi, \tau)$ represented as sequence
of state and action tuples $\{(x^j, u^j)\}_{j=0}^{n}$ with $x^n = g$.
Each state in this solution $x^{j+1} = \Gamma(x^j, u^j)$ is the physical outcome of
the previous action applied to the previous state. If the algorithm failed to add a goal within
$n_\text{max}$ iterations, a special value $\emptyset$ is returned to indicate failure.

\subsubsection{Sampling}
Each iteration starts with the $\textsc{SampleSlice}$ function, which
samples a slice by sampling an object arrangement. This is done by either
uniformly sampling the space of object arrangements $\mathcal{C}_{1:m}$, or
with some probability $p_g$, by sampling the set of goal arrangements in $\mathcal{G}$.
Similarly, the function $\textsc{RandomChoice}$ samples an index $t \in \{0, 1, \ldots, m\}$
with bias towards either the robot $t = 0$ or a target object $t \in \mathbf{T}$.

\subsubsection{Selecting Nearest Neighbors}
To select the nearest explored slice $s^j \in \mathcal{S}$, we require a distance function $d_s$.
Measuring this distance, however, is challenging. Ideally, we would apply
a distance function that expresses a minimal cost that it takes the robot to move all objects
from one slice to the other. Such a cost, however, is generally not available, as
it would require to take all robot motions needed to clear occlusions as well as
simultaneous object manipulation into account.
Instead, we opt to approximate this through the distance function
\begin{equation}
        d_s(s^a, s^b) = \sum_{i=1}^m d_{\mathcal{C}_i}(s^a_i, s^b_i).
        \label{eq:slice_distance}
\end{equation}
that sums the distances of the individual object states.

\subsubsection{Extending the Tree}
In the $\textsc{ExtendTree}$ function, \algref{algo:extend_tree},
the search tree is either extended using a targeted extension strategy
or with some small probably $p_{rand}$ by a random action.
When following our strategy, it first uses a steering function for the robot to compute an action $u_0$
that attempts to move the robot from its current state $x_0^j$ towards
the sampled state $x^d_0$. This action is then forward propagated in the subroutine $\textsc{ExtendStep}$
using the physics model $\Gamma$. If this action leads to a valid state, the
$\textsc{ExtendStep}$ function adds the resulting state to the tree $\mathcal{T}$ and updates
$\mathcal{S}$ accordingly. Note that although
this action is intended to move only the robot, it may result in collision and thus
also move any of the $m$ movable objects. The $\textsc{ExtendStep}$ function therefore
returns the resulting composite state $x^c = \Gamma(x^j, u_0)$.

If $t$ is not the robot, $x^d_0$ is a manipulation state provided by our generator to push object $t$.
Hence, the algorithm next queries the policy $\pi_\theta$ in $\textsc{QueryPolicy}$ to achieve the desired push.
The returned action is then again passed to the $\textsc{ExtendStep}$ function and $\mathcal{T}$ and
$\mathcal{S}$ are updated accordingly.

\subsubsection{Physics Propagation}
Given a state $x \in \mathcal{C}_{0:m}^\text{free}$
the $\textsc{ExtendStep}$ function forward propagates an action $u$ through the
physics model $\Gamma$. An action is considered invalid, if the scene fails
to come to rest within some time limit, i.e.\ $\Gamma(x, u) = \bot$, or
if the action leads to collisions of either the robot or any movable object with static obstacles.
Collisions between movable objects, in contrast, are allowed and modeled accordingly.
If the propagation is valid and $x' = \Gamma(x, u)$ is within bounds, the search
tree $\mathcal{T}$ and $\mathcal{S}$ are updated accordingly.

\begin{algorithm}[t]
    \algsetup{linenosize=\tiny}
    \small
    \SetFuncSty{textsc}
    \DontPrintSemicolon
    \SetKw{Not}{not}
    \SetKw{Is}{is}
    \SetKwInOut{Constants}{Constants}
    \KwIn{Object identifier $t$,
        current tree node $x^j \in \mathcal{T}$ to extend,
        state to extend towards $x^d \in \mathcal{C}_{0:m}$,
        search tree $\mathcal{T}$, explored slices $\mathcal{S}$}
    \Constants{$p_{rand} \in [0,1]$}
    \KwOut{Updated $\mathcal{T}$ and $\mathcal{S}$}
    \SetKwFunction{Steer}{Steer}
    \SetKwFunction{QueryAction}{QueryPolicy}
    \SetKwFunction{ExtendStep}{ExtendStep}
    \SetKwFunction{Uniform}{Uniform}
        \If{\Uniform{$[0,1]$} $\leq p_{rand}$} {
                $u \gets$ \Uniform{$U$} \;
                $\bot, \bot, \mathcal{T}, \mathcal{S} \gets $\ExtendStep{$x^j, u, \mathcal{T}, \mathcal{S}$} \;
        } \Else {
            $u_0 \gets$ \Steer{$x_0^j, x_0^d$} \;
            $\textit{valid}, x^c, \mathcal{T}, \mathcal{S} \gets $\ExtendStep{$x^j, u_0, \mathcal{T}, \mathcal{S}$} \;
            \If{\textbf{not} $\text{valid}$ \textbf{or} $t \neq r$} {
               \Return $\mathcal{T}, \mathcal{S}$ \;
            }
            $u \gets $ \QueryAction{$x^c_0, x_t^c, x_t^d$} \;
            $\bot, \bot, \mathcal{T}, \mathcal{S} \gets $\ExtendStep{$x^c, u, \mathcal{T}, \mathcal{S}$} \;
        }
        \Return $\mathcal{T}, \mathcal{S}$
\caption{The \textsc{ExtendTree} function}
\label{algo:extend_tree}
\end{algorithm}

\subsubsection{Extracting and Shortcutting the Path}
If the algorithm succeeds at adding a goal state to the tree,
the function $\textsc{ExtractAndShortcut}$ extracts the solution
$(x^0, u^0), \ldots, (x^n, \bot)$ with $x^n \in \mathcal{G}$ from $\mathcal{T}$.
Due to the randomization of the algorithm, this solution may contain
multiple actions that are not required. We shortcut these using
the shortcut algorithm presented by King et.\ al \cite{King2015}. The algorithm selects
two pairs $(x^i, u^i), (x^j, u^j)$ and replaces $u^i$ with an action moving the robot from
state $x^i_0$ to $x^j_0$. Thereafter, it forward propagates this new action and all remaining actions $u^k$ with
$j \leq k < n$ through the physics model to probe whether the goal can still be achieved.
If this is the case and the resulting solution is shorter in terms of execution time,
the old solution is replaced by the new one. This process is repeated until either all pairs have been
attempted or a timeout occurred. The pairs can be selected in different orders, although uniform
random sampling performed best in our experiments.

\subsection{Learning the Policy and the Generator}
\label{sec:method:oracle}
The $\textsc{QueryGenerator}$ and $\textsc{QueryPolicy}$
functions provide the planner with a manipulation state and an action
to transport a single object $i$ from a state $x_i$ to some desired state $x_i^d$.
As illustrated in \secref{sec:concept}, there might be multiple such manipulation states that the
generator could sample. Similarly, in general there are several actions
that the policy could select to achieve the \textit{transfer}.
We can represent these two sets implicitly through two families of distributions
\begin{align}
    &p(x_0|x_i^d, x_{1:m}, \nu_{0:m}, \mathcal{O}), \label{eq:generator_family} \\
    &p(u | x_i^d, x_{0:m}, \nu_{0:m}, \mathcal{O}) \label{eq:policy_family}
\end{align}
that place all probability mass on the manipulation states and actions for
the desired object transport. Both types of distributions are conditioned
on the state of all objects $x_{1:m}$, physical properties of the objects and the robot $\nu_{0:m}$,
and all static obstacles~$\mathcal{O}$. The action distributions in \eqref{eq:policy_family} are additionally
conditioned on the state of the robot $x_0$.

With this representation, learning the generator and the policy could be achieved by learning parameters
$\theta$ and $\phi$ that characterize instances from these families of distributions.
Once learned, the generator and the policy could then produce the desired samples by
sampling from these distributions. Learning $\theta$ and $\phi$ for this
general case, however, is very difficult. The learned distributions would need
to be conditioned on an arbitrary number of movable objects $m$ as well
as all possible static obstacle configurations $\mathcal{O}$.

Therefore, instead, we simplify the problem to learning distributions of the form
\begin{align}
    &p(x_0|x_i^d, x_{i}, \nu_{i}), \label{eq:simplified_generator_family} \\
    &p(u | x_i^d, x_0, x_{i}, \nu_{i}) \label{eq:simplified_policy_family}.
\end{align}
In other words, we choose to learn the generator and the policy in an obstacle-free world
containing only the robot and a single object. We leave it to the planning algorithm to find
manipulation states and actions that are feasible in the full problem. Also, we learn
both policy and generator for a single robot at a time. Applying the planner to a different robot
requires training a different policy and generator. For brevity, we
thus omit the dependency on the robot's physical properties $\nu_0$.

In the following, we first present how we learn the policy, i.e.\ an instance of an action distribution,
from data generated offline. Thereafter, we derive an instance of a manipulation state distribution from
this policy and present how the generator can be trained to efficiently sample
from this distribution within the planner.

\subsubsection{Learning the Policy}
\label{sec:method:learning:policy}

The parameter $\theta$ that characterizes the policy
can be found by solving the problem
\begin{equation}
    \label{eq:general_policy_expectation}
        \min_{\theta} \mathbb{E}_{x_0, x_i^d, x_i, \nu_i}\Bigl[\mathbb{E}_{x_i', u}[d_{\mathcal{C}_i}(x_i',x_i^d)       ~|~x_0, x_i^d, x_i, \nu_i, \theta] \Bigr]
\end{equation}
over some training distributions for $x_0, x_i^d, x_{i}, \nu_{i}$.
The inner expected value
\begin{multline}
    \label{eq:general_policy_integral}
    \mathbb{E}_{x_i', u}[d_{\mathcal{C}_i}(x_i',x_i^d)~|~x_0, x_i^d, x_i, \nu_i, \theta] = \\
    \begin{aligned}
        \iint d_{\mathcal{C}_i}(x_i',x_i^d)\,&p(x_i'|u, x_0, x_i, \nu_i)\,dx_i'\\ &p(u | x_i^d, x_0, x_i, \nu_i, \theta)\,\,du
    \end{aligned}
\end{multline}
is the expected distance between a desired object state $x_i^d$ and the actual
state $x_i'$ that the policy parameterized by $\theta$ achieves, given it is
executed when the robot is in state $x_0$ and the object with properties $\nu_i$ is in state $x_i$.
The successor state distribution $p(x_i'|u, x_{0}, x_i, \nu_{i})$ describes
the physical outcome of an action. Although we could model this using the
deterministic physics model $\Gamma$, it proves useful to apply a
non-deterministic model for learning the policy. We
will motivate this shortly.

We simplify the expression in \eqref{eq:general_policy_integral} by
choosing our policy to be a deterministic function
$u_\theta \coloneqq \pi_\theta(x_0, x_i, x_i^d, \nu_i)$.
In other words, we enforce the action distribution to be a Dirac delta function
$p(u|x_i^d, x_0, x_i, \nu_i, \theta)~=~\delta(u~-~\pi_\theta(x_0, x_i, x_i^d, \nu_i)).$
With this, the expected distance in \eqref{eq:general_policy_integral} reduces to
\begin{equation}\label{eq:energy_function1}
    \begin{aligned}
        L_\theta(x_0, x_i, x_i^d, \nu_i) \coloneqq&\int d(x_i',x_i^d)\,p(x_i'|u_\theta, x_0, x_i,\nu_i)\,dx_i' \\
        =&~\mathbb{E}_{x_i'}[d_{\mathcal{C}_i}(x_i',x_i^d)~|~u_\theta, x_0, x_i, \nu_i]
    \end{aligned}
\end{equation}
and our problem of learning the policy becomes
\begin{equation}
    \label{eq:simplified_policy_expectation}
    \min_{\theta} \mathbb{E}_{x_0, x_i^d, x_i, \nu_i}\Bigl[ L_\theta(x_0, x_i, x_i^d, \nu_i) \Bigr].
\end{equation}

To solve this problem efficiently, we need to have access to the gradient of
the loss $L_\theta(x_0, x_i, x_i^d, \nu_i)$ w.r.t. $\theta$.  To acquire this, we choose
the distance function on $\mathcal{C}_i \subseteq SE(2)$ to be of the form
$d_{\mathcal{C}_i}(x_i', x_i^d)=\sum_{j=1}^3\sigma_j(x_{ij}'~-~x_{ij}^d)^2$
with positive weights $\{\sigma_j\}_{j=1}^3$. The loss then becomes
\begin{equation}
    \begin{aligned}
        L_\theta(x_0, x_i, x_i^d, \nu_i)&=
        \sum_{j=1}^3 \sigma_j \mathbb{E}_{x_i' }\left[ (x_{ij}' - x_{ij}^d)^2~|~u_\theta, x_0, x_i, \nu_i \right],
    \end{aligned}
\end{equation}
where $x_{i0}, x_{i1} \in \mathbb{R}$ denote the object's position and $x_{i2}~\in~SO(2)$ its orientation.
Expanding the square and rewriting yields
\begin{multline}\label{eq:variance_mean_decomp}
        L_\theta(x_0, x_i, x_i^d, \nu_i) = \\
        \sum_{j=1}^3 \sigma_j \left( \text{Var}_{x_i'} (x_{ij}' ) + \left(\mathbb{E}_{x_i'}\left[ x_{ij}'\right] - x_{ij}^d\right)^2 \right),
\end{multline}
where both variance and expected value are conditioned on $x_0, x_i, \nu_i, u_\theta$.

This means we can express the loss $L_\theta$ as a function of forward models
\begin{equation}\label{eq:forward_mean}
    f_\mu(x_0, x_i, \nu_i, u) = \mathbb{E}_{x_i'}\left[x_i' | x_0, x_i, \nu_i, u \right]
\end{equation}
and
\begin{equation}\label{eq:forward_var}
    f_{\sigma^2}(x_0, x_i, \nu_i, u) = \text{Var}_{x_i'}(x_i' | x_0, x_i, \nu_i, u ).
\end{equation}
In particular, if these forward models are differentiable w.r.t. to the action argument $u$, we can
apply the chain rule in \eqref{eq:variance_mean_decomp} to obtain the gradient of $L_\theta$
w.r.t. $\theta$.
With this deterministic policy gradient~\cite{silver2014deterministic}
we can then learn the policy by solving~\eqref{eq:simplified_policy_expectation}.

We can obtain differentiable models $f_\mu$ and $f_{\sigma^2}$ through supervised learning from a
data set $\{(x_0^j, x_i^j, \nu_i^j, u^j, (x_i')^j)~|~j\in~\mathbb{N}\}$ of robot-object interactions.
We generate this data set offline using the physics model $\Gamma$.
Learning the models $f_\mu$ and $f_{\sigma^2}$ in this way has the advantage that
we are free in choosing any physics model $\Gamma$ for generating the data.
In particular, this allows us to choose models $\Gamma$ for which gradients w.r.t. $u$
are not available, as it is the case for many rigid body physics simulators.

\eqref{eq:variance_mean_decomp} also provides insight into why we choose
to apply a non-deterministic physics model for learning the policy.
If we applied a deterministic model, i.e.\ the successor state distribution
$p(x_i'|u, x_{0}, x_i, \nu_{i})$ had zero-variance, the variance term
in \eqref{eq:variance_mean_decomp} would disappear. Hence, our
policy could select any action $u \in U$ for a state $(x_0, x_i)$
that transports the object as close as possible to $x_i^d$.
Not all of these actions, however, are equally desirable in practice.
For instance, a robotic hand pushing an object with the tip of its finger
only works reliably if the object behaves exactly as predicted by the physics model.
Pushing the object with the palm of the hand, in contrast, is
more reliable, since the object is trapped between the fingers.
Thus, to raise preference for the policy to learn reliable actions,
we add observation noise to the object state $x_i$,
and the object's mass and friction parameters in $\nu_i$ when generating our data set.
Since the loss in \eqref{eq:variance_mean_decomp} is minimized by actions that achieve low
variance in successor state, this will result in our policy selecting actions
that achieve the desired push even under uncertainty in the modified physics
parameters.

\subsubsection{Acquiring a Manipulation State Distribution}
Given the trained policy, we derive a manipulation state distribution parameterized by $\phi$ by solving
\begin{equation}
    \min_\phi \mathbb{E}_{x_i^d, x_i, \nu_i} \Bigl[\mathbb{E}_{x_0}\left[L_\theta(x_0, x_i, \nu_i, x_i^d)~|~x_i, \nu_i, x_i^d, \phi \right]\Bigr]
\end{equation}
over training distributions for $x_i^d, x_i, \nu_i$.
Here, the inner expected value
\begin{multline}
    \mathbb{E}_{x_0}\left[L_\theta(x_0, x_i, \nu_i, x_i^d)~|~x_i, \nu_i, x_i^d,\phi \right] =\\
    \int_{\mathcal{C}_0} L_\theta(x_0, x_i, \nu_i, x_i^d)\,p(x_0 | x_i, x_i^d, \nu_i, \phi)\,dx_0
\end{multline}
is the expected loss of the trained policy if we choose initial robot states according to
the distribution parameterized by $\phi$.

In principle, this expected value would be minimized by a Dirac delta function that places all its probability
mass at a robot state that minimizes the loss $L_\theta$ given the arguments $x_i, x_i^d, \nu_i$.
This, however, is undesirable. The loss $L_\theta$ is defined on a simplified problem that
only considers a single object. The state that minimizes $L_\theta$ may not be
feasible in the full problem with obstacles that is addressed by the planner.
Hence, instead, we opt for a wider distribution that provides the planner with
more diverse manipulation state samples. Namely, we choose $\phi$ to describe
an energy-based distribution of the form
\begin{equation}\label{eq:energy_distribution}
    p(x_0|x_i^d, x_i, \nu_i) = \frac{1}{Z}e^{-\lambda L_\theta(x_0, x_i, x_i^d, \nu_i)},
\end{equation}
where $Z$ is a normalization constant and  $\lambda~\in~\mathbb{R}^+$ a manually
selected parameter.

We choose this distribution since its modes are placed on the robot states
that allow pushing the object to the desired state in the obstacle-free world.
By letting $\lambda \rightarrow \infty$, we get a distribution with all its
probability mass placed on the minima of $L_\theta$ w.r.t. to the robot state.
Vice versa choosing small $\lambda$, we get a distribution with wider support,
which increases the chance of the distribution to cover the feasible manipulation states
of the full problem.

\subsubsection{Learning a Generative Model}
The forward models $f_\mu$ and $f_{\sigma^2}$ allow us to compute
$L_\theta$, and thus the density function in \eqref{eq:energy_distribution} efficiently.
For the $\textsc{QueryGenerator}$ routine in the planner, however, we need to produce
samples of robot states following this distribution.
We can acquire these using the Markov Chain Monte Carlo (MCMC) method:
\begin{equation}
    x_0 \sim p_{\text{MCMC}}(x_0|x_i, x_i^d, \nu_i).
\end{equation}
This, however, takes several iterations for the distribution to
burn in and multiple forward passes through the learned forward models, making this
computationally too expensive to be used inside the planner. Hence, instead,
we generate a set of samples using this method offline, and train a (conditional)
generative adversarial network (GAN)~\cite{goodfellow2014generative}
to mimic this distribution online.
This way we can generate samples from the distribution with a single
forward pass through a neural network.

A GAN consists of two parts: a generator $G_{\varphi_G}$, and a discriminator $D_{\varphi_D}$.
Both are neural networks parameterized by $\varphi_G$ and $\varphi_D$ respectively.
The generator $G_{\varphi_G}(x_i, x_i^d, \nu_i, z)$ is a function
that maps a random real-valued sample $z \sim \mathcal{N}(\mathbf{0}, \mathbf{I})$ and its arguments
$x_i, x_i^d, \nu_i$ to a robot state sample. The discriminator outputs the probability that a
robot state sample $x_0$ is drawn from the \textit{true} distribution of manipulation states
$p_{\text{MCMC}}(x_0|x_i, x_i^d, \nu_i)$ that we obtain using MCMC.
Both models are trained by solving the following problem:
\begin{equation}
        \label{eq:gan_objective}
        \min_{\varphi_G} \max_{\varphi_D} V(\varphi_D, \varphi_G)
\end{equation}
with
\begin{multline}
        V(\varphi_D, \varphi_G) =
        \mathbb{E}_{x_0 \sim p_\text{MCMC}}\left[ \log(D_{\varphi_D}(x_0, x_i, x_i^d, \nu_i)) \right] + \\
        \mathbb{E}_{z}\left[\log(1 -  D_{\varphi_D}(G_{\varphi_G}(z, x_i, x_i^d, \nu_i), x_i, x_i^d, \nu_i) )\right]
\end{multline}
On the one hand, this objective trains the discriminator to distinguish between
true samples from $p_{\text{MCMC}}$ and samples generated by the generator.
On the other hand, it optimizes the generator to produce samples such that the discriminator
can not distinguish these generated samples from the true samples from $p_{\text{MCMC}}$.
Once learned the generator then serves as an efficient state sampler in the \textsc{QueryGenerator} function.

\subsubsection{Summary}
To summarize, the procedure to learn the deterministic policy $\pi_\theta$ and the generator
$G_\varphi$ consists of four steps illustrated in \figref{fig:system_overview} on the left.
First, we generate a data set of robot-object interactions using the physics model $\Gamma$.
Second, we train two forward models $f_\mu$ and $f_{\sigma^2}$ from this data set under observation
noise of the physical properties of the training objects. Third,
we use these forward models to compute the loss in \eqref{eq:variance_mean_decomp}
and train the policy $\pi_\theta$.
Fourth, we define an approximate manipulation state distribution
according to \eqref{eq:energy_distribution} and train the generator $G_\varphi$ to
imitate samples following this distribution.

    \begin{figure*}[ht]
        \setlength{\tabcolsep}{0pt}
        \begin{tabular}{|c|c|c|c|c|}
                \hline
                \begin{subfigure}[b]{0.198\textwidth}
                        \includegraphics[width=\textwidth]{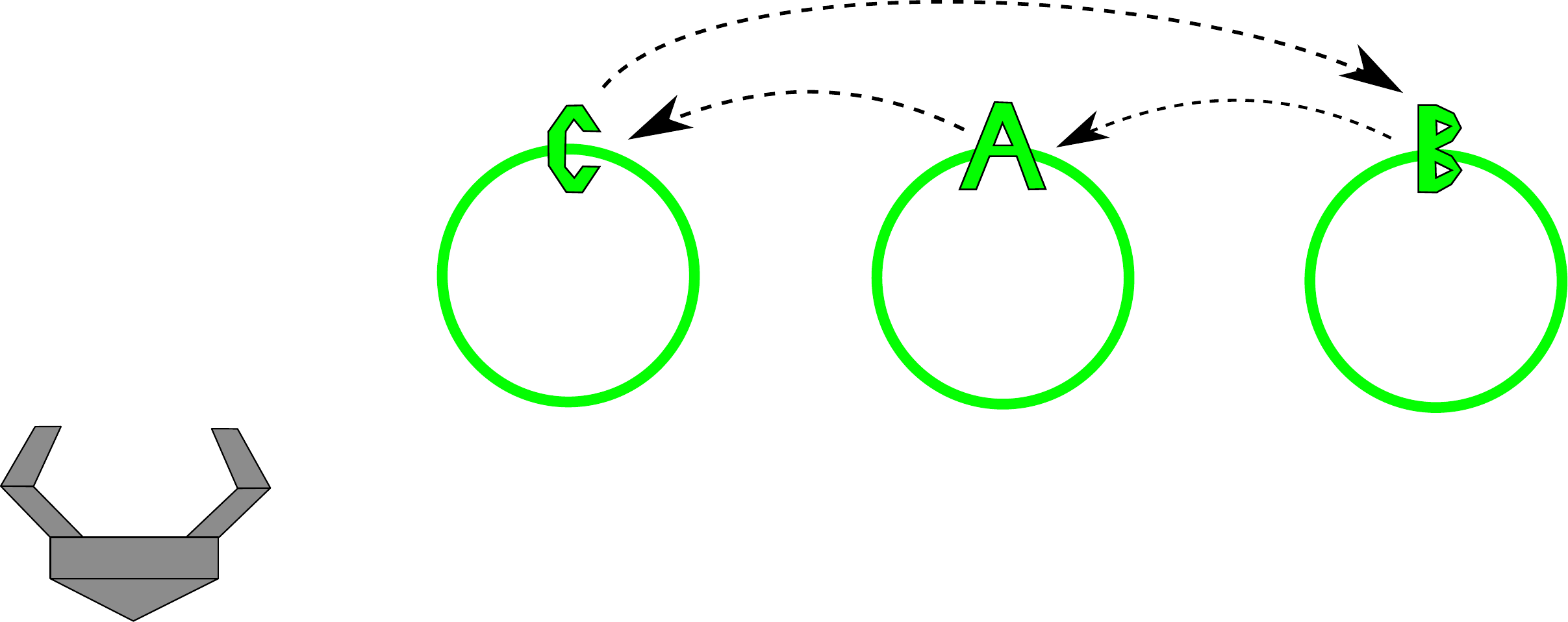}
                   \caption{ABC Problem}
                   \label{fig:results:abc_scene}
                \end{subfigure}
                &
                \begin{subfigure}[b]{0.198\textwidth}
                        \includegraphics[width=\textwidth]{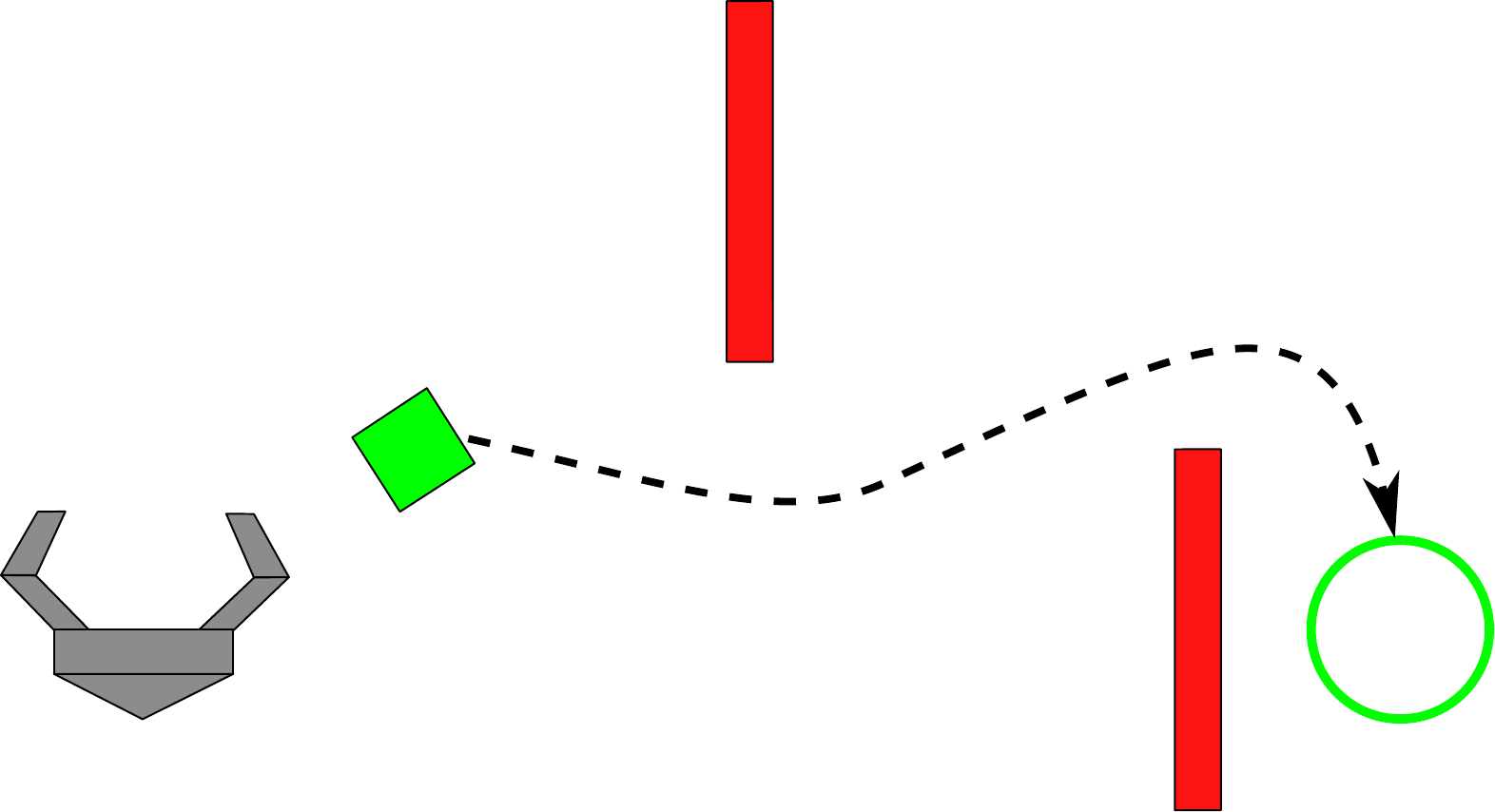}
                        \caption{Slalom and Slippery Slalom}
                        \label{fig:results:slalom_scene}
                \end{subfigure}
                &
                \begin{subfigure}[b]{0.198\textwidth}
                        \center
                        \includegraphics[width=0.95\textwidth]{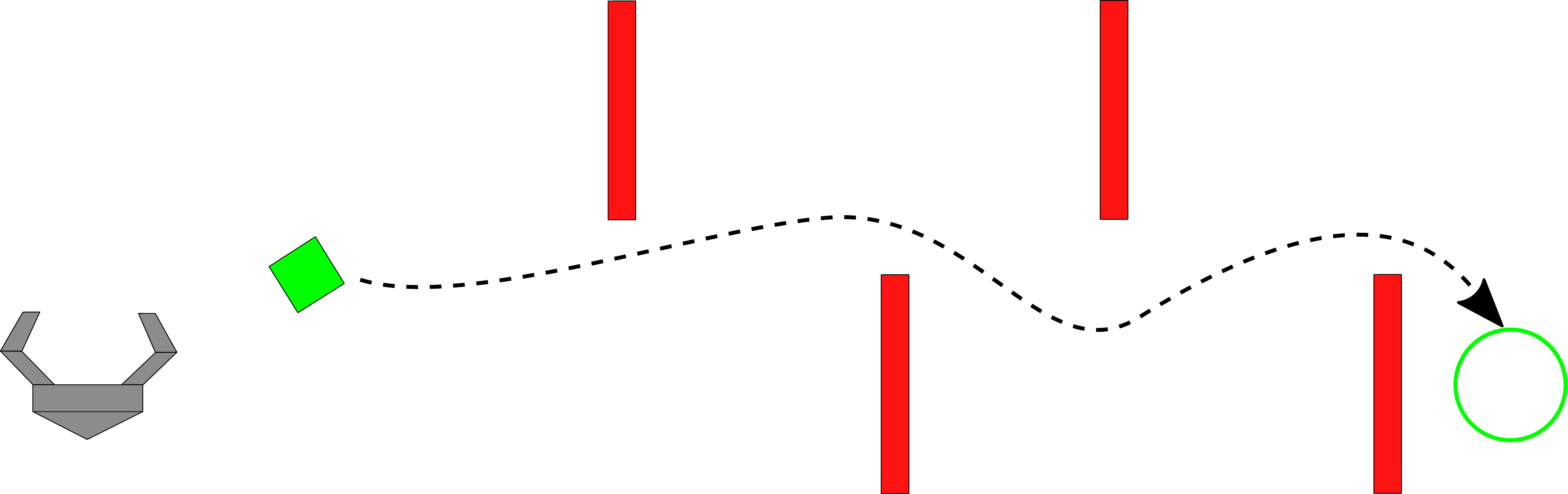}
                   \caption{Long Slalom}
                   \label{fig:results:long_slalom_scene}
                \end{subfigure}
                &
                \begin{subfigure}[b]{0.198\textwidth}
                        \center
                        \includegraphics[width=0.95\textwidth]{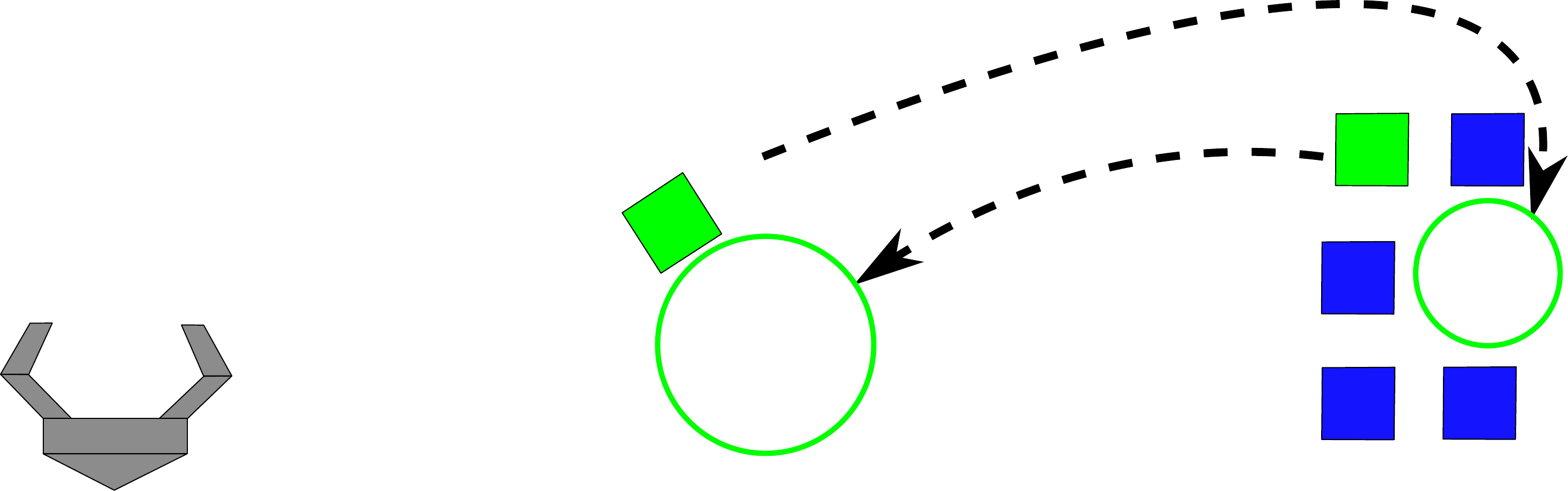}
                        \caption{Movable Cage}
                        \label{fig:results:obstacle_wall_scene}
                \end{subfigure}
                &
                \begin{subfigure}[b]{0.198\textwidth}
                        \includegraphics[width=\textwidth]{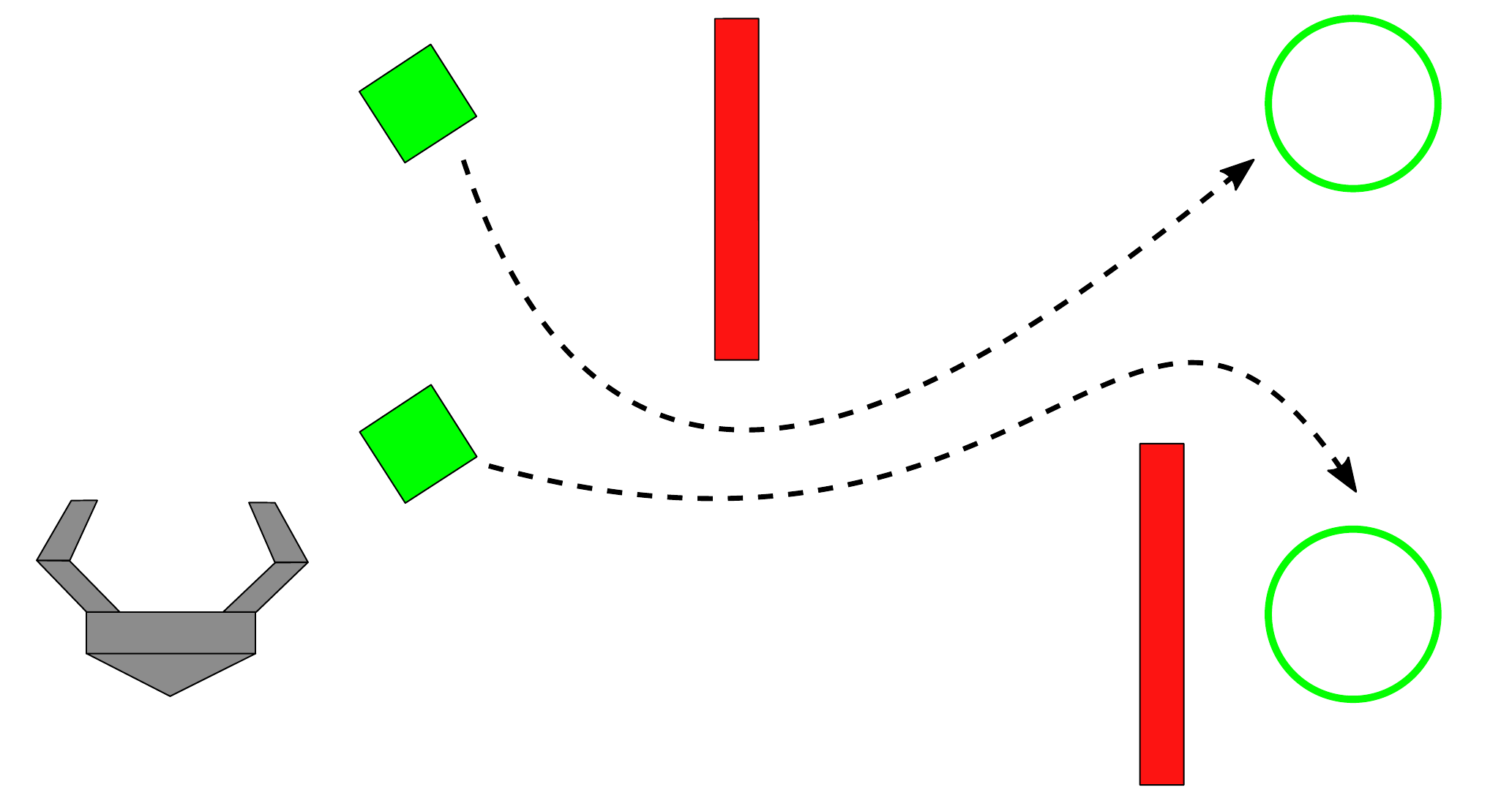}
                        \caption{Dual Slalom}
                        \label{fig:results:dual_slalom}
                \end{subfigure}
                \\
                \hline
        \end{tabular}
        \caption{
                Test scenes for our experiments. These are chosen to test a
                variety of features: differing object properties, with and
                without static obstacles, movable obstacles, rearranging
                single/multiple objects. The scenes Slalom and Slippery Slalom
                differ in the friction coefficients and mass of the object. In
                the ABC Problem the objects differ in shape, mass and friction
                from each other. Green objects are targets $\mathbf{T}$, red
                objects are static obstacles and blue objects are movable
                objects that are allowed to be moved anywhere, but outside of
                the planning scene. Consult our online supplementary material
                for videos of example solutions.
        }
        \label{fig:scenes}
\end{figure*}

\section{Experiments}
\label{sec:experiments}
We implemented and evaluated our approach for robots endowed with
configuration spaces in $SE(2)$, i.e.\ $\mathcal{C}_0 \subseteq SE(2)$.
Evaluating the approach for robots with higher dimensional configuration spaces
raises additional challenges that we will discuss in \secref{sec:discussion}.
For simplicity, we additionally limit our evaluation to holonomic robots. We note though
that our approach is not limited to such robots as long as we can efficiently compute a steering function.
Unless stated otherwise, we apply in our experiments a
planar robot with the geometry of a robotic gripper, as shown in \figref{fig:scenes}.
We choose this geometry because it has interesting properties for the system
to learn about.

The action space of all robots in our experiments is the space of bounded
translational and rotational velocities applied for some bounded duration:
$$U = \{(v_x, v_y) \in \mathbb{R}^2~|~v_x^2 + v_y^2 \leq \hat{v}\} \times [-\hat{\omega}, \hat{\omega}] \times [0, \hat{t}\,]$$
with bounds $\hat{v}, \hat{\omega}$ and $\hat{t}$.
As the robot is required to be at rest after each action,
the actions follow a ramp velocity profile with linear acceleration and deceleration phases
similar to \cite{Haustein2015}. Accordingly, the robot's steering
function computes an action that moves the robot on a straight line between two poses
ignoring obstacles.

We first present how the robot state generator and policy
for this type of robot can be learned. Thereafter, we evaluate
these learned models and their performance within the planning algorithm.
Additionally, we evaluate how our different design choices in the algorithm affect its
planning efficiency.

\subsection{Learning Generator and Policy for $SE(2)$-Robots}
\begin{figure*}[!ht]
    \centering
    \includegraphics[width=0.9\textwidth]{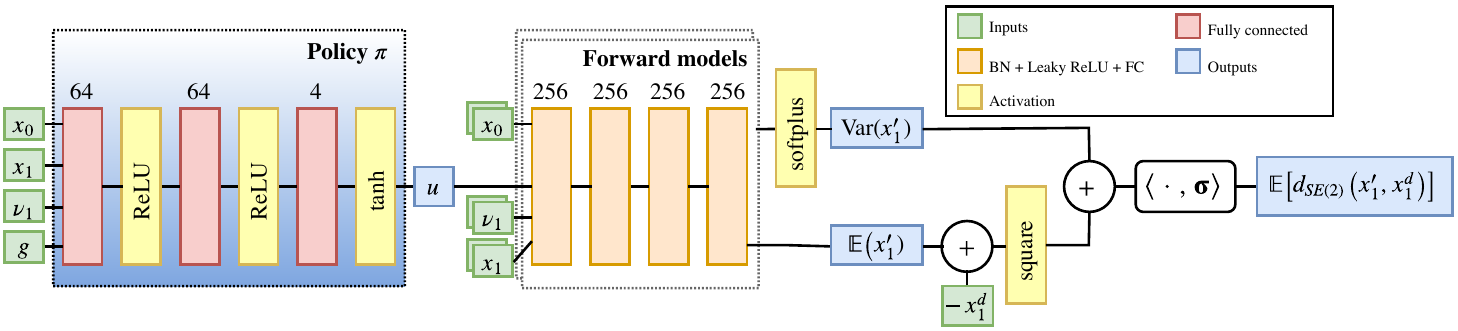}
    \caption{Architecture of forward models and the policy. The forward models are
             first trained on input tuples ($x=(x_0, x_1), u, x'=(x_0', x_1'), \nu$).
             The policy can then be trained by minimizing the rightmost output for
             randomly sampled desired object states $x_1^d$, and
             back-propagating through the forward models. The forward models are using
             batch normalization (BN) \cite{ioffe2015batch} and leaky ReLUs
         \cite{maas2013rectifier}.}
    \label{fig:policy_overview}
\end{figure*}
\subsubsection{Data Generation}
The data set used for training the forward models $f_\mu$ and $f_{\sigma^2}$
consists of tuples $(x_0, x_i, u, x_0', x_i', \nu_i)$.
In order for both the generator and the policy to be applicable to different object types,
we generate this data for objects $i$ with different shapes, sizes, mass and friction coefficients.
The parameters $\nu_i$ describing these objects for the generator and policy contain
the width and height of the minimal bounding box, the mass and inertia, and the friction
coefficient between ground and object.
We collect the data by first randomly sampling a robot-object state
$x \in \mathcal{C}_{0:1}$ and an action $u \in U$. We then forward propagate these
through a stochastic physics model, which we acquire from $\Gamma$
as described in \secref{sec:method:learning:policy}.
If the resulting state is valid, i.e.\ the object comes to rest within a time limit $T_{\text{max}} = 8s$,
we add the tuple to our data set.
Lastly, since the robot's configuration space is also in $SE(2)$,
we can exploit redundancy in our learning problem.
That is, the results of the physical interaction between the robot and the object are translationally
and rotationally invariant. Therefore, we transform all states into a common reference frame
such that the robot state is placed at the origin, $x_0 = \mathbf{0}$.

\subsubsection{Forward Models}
Both forward models $f_{\sigma^2}$ and $f_\mu$ are learned with
neural networks shown in \figref{fig:policy_overview}.
The optimization is done using Adam \cite{kingma2014adam} in mini-batches of size 256
for \numprint{64000} steps. The final models are chosen based on validation scores on a held-out validation set.
We train both models by maximizing the log-likelihood of a multivariate
Gaussian, since the maximum likelihood estimate of normally distributed
variables is exactly the mean and the variance. Since we are only interested
in the variances, we let the covariance matrix be a diagonal matrix.

\subsubsection{Policy}
Based on the learned forward models we can learn the policy $\pi_\theta$
by minimizing \eqref{eq:variance_mean_decomp}.
The architecture of the policy is shown in \figref{fig:policy_overview}.
The policy is trained in batches from the same data as the forward models.
We augment the data tuples $(x_0, x_i, \nu_i, u, x')$ with object goal states $x_{i}^d$ by randomly sampling
from a Gaussian over the true successor state $x_i'$ observed in the data set.
The policy is trained for \numprint{40000} steps with the same optimizer and
mini-batch size as the forward models.

\begin{figure*}[!ht]
    \centering
    \includegraphics[width=0.9\textwidth]{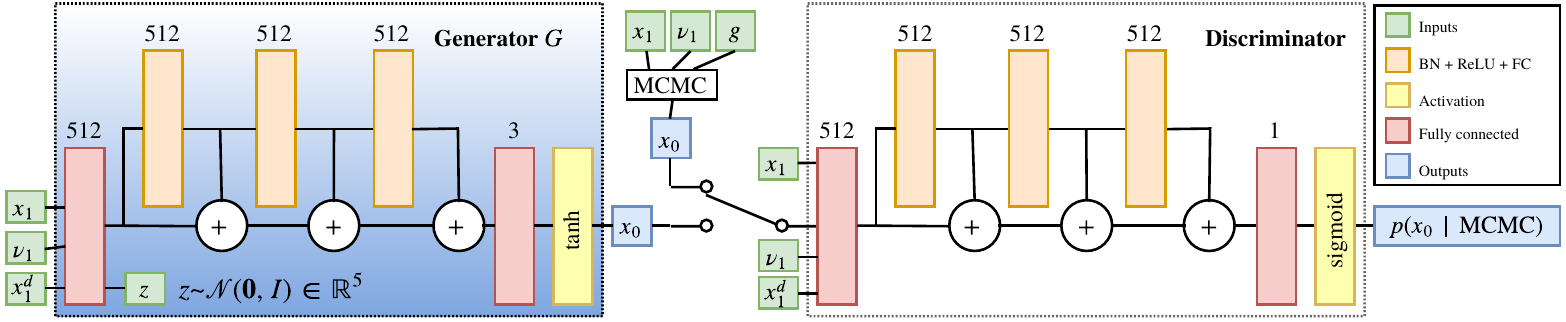}
    \caption{GAN architecture. The generator is trained to mimic the samples
             that a slower Markov Chain Monte Carlo algorithm provides. Samples
             can then instead be obtained by a single forward pass of the generator network.}
    \label{fig:gan_overview}
\end{figure*}

\subsubsection{Generative Model}\label{sec:feasible_state_sampling}
The GAN is trained from MCMC samples of robot states.
To generate these we first draw samples of initial object
and successor states $x_i, x_i'$ from our data set.
These samples are then translated to object frame, i.e.\ such that $x_i = \mathbf{0}$.
This simplifies the application of the transition kernel to produce proposal states for the MCMC method.
The initial proposals of robot states are drawn from a Gaussian distribution $\mathcal{N}(0, 0.01^2)$
for the positions, and a uniform distribution $U(-\pi, \pi)$ for the rotation.
Then, new states are proposed by adding Gaussian noise $\mathcal{N}(0, 0.05^2)$ to the position,
and Gaussian noise $\mathcal{N}(0, 0.5^2)$ to the rotation. Given a current robot state sample $x_0$,
a new sample $x_0'$ is accepted by the MCMC method with probability $\alpha$, defined as:
\begin{equation}
    \alpha = \min(1, e^{\lambda \left(L_\theta(x_0,x_i,x_i',\nu_i) - L_\theta(x_0',x_i,x_i',\nu_i)\right)})
\end{equation}
Here, $\lambda$ is a temperature parameter that we set in our experiments set to $128$.
We used a burn-in of 100 iterations and thereafter added the following $300$ samples to a data set
for training the GAN. In total, we collected $2 \times 10^6$ MCMC samples for each GAN training.

The architecture of the GAN is shown in \figref{fig:gan_overview}.
In practice, the GAN objective from \eqref{eq:gan_objective} is split up into the following
three objectives:
\begin{equation}
    \max_{\varphi_G}  \mathbb{E}_{z}\left[\log(D_{\varphi_D}(G_{\varphi_G}(z, x_i, x_i^d, \nu_i)) )\right]
\end{equation}
\begin{equation}
    \max_{\varphi_D}  \mathbb{E}_{x_0 \sim p_\text{MCMC}}\left[ \log(D_{\varphi_D}(x_0, x_i, x_i^d, \nu_i)) \right]
\end{equation}
\begin{equation}
    \min_{\varphi_D} \mathbb{E}_{z}\left[\log(D_{\varphi_D}(G_{\varphi_G}(z, x_i, x_i^d, \nu_i)) )\right]
    \label{eq:discriminator_fake_loss}
\end{equation}
All three objectives are optimized in mini-batches of size 256 with RMSProp. To stabilize training,
the loss in \eqref{eq:discriminator_fake_loss} is trained with probability $0.5$ on a batch of generated
samples from a replay buffer, and otherwise on a recent batch from the generator.
The discriminator is regularized by adding the square of the logits
to the loss function. The training is run for \numprint{100000} iterations.

\subsection{Baselines for Evaluation}
To evaluate our algorithm and our learned models, we define several baselines.

\subsubsection{Simple Pushing Heuristic}
To evaluate the learned generator and policy,
we define a simple generator and policy that share similarities with typical
pushing primitives applied in prior rearrangement planning works.
To sample a manipulation state, we sample
a distance $r \sim \mathcal{N}(\mu_r, \sigma_r^2)$ and an angle
$\theta \sim \mathcal{N}(0, \sigma_\theta^2)$ for some manually specified $\mu_r, \sigma_r$
and $\sigma_\theta$. The robot is then placed next to the target object at the position
$\mathbf{x}_0 = \mathbf{x}_1 - r R_\theta \frac{\mathbf{x}_1^d - \mathbf{x}_1}{\|\mathbf{x}_1^d - \mathbf{x}_1\|}$,
where $\mathbf{x}_1,\mathbf{x}_1^d \in \mathbb{R}^2$ are the current and the desired object position,
and $R_\theta \in SO(2)$ a rotation matrix by angle $\theta$. The robot's orientation
is selected such that its palm faces in the pushing direction. From this robot pose
our simple policy steers the robot in a straight line to a position that is offset
by the dimensions of the object from the target position $\mathbf{x}_1^d$.

\subsubsection{HybridActionRRT}
To evaluate our planning algorithm, we compare it to King et al.'s~\cite{King2016}
HybridActionRRT algorithm. In brief, this algorithm differs from ours in that it
does not structure $\mathcal{C}_{0:m}$ in slices and applies a different strategy
for extending the search tree. The search tree is grown by first randomly sampling a full
state $x^d \in \mathcal{C}_{0:m}$ and then extending the tree from the state that is closest
to the sample $x^d$. Herein, \textit{closest} is defined by a distance function on $\mathcal{C}_{0:m}$ that
takes the states of all objects and the robot into account. In our experiments, we apply a distance
function similar to \eqref{eq:slice_distance} with an additional equally weighted
term for the distance in robot state.

The $\textsc{ExtendTree}$ function of this algorithm samples $k$ action sequences,
where each sequence is either with probability $p_\mathit{rand}$
sampled uniformly at random or a noisy sample of actions provided by a manipulation primitive.
Thereafter, all $k$ sequences are forward propagated through the physics model
and the one that results in a state closest to the random sample $x^d$ is added to the tree.
We equip this algorithm with two manipulation primitives:
\begin{itemize}
   \item[1] \textit{Transit}: The robot is steered on a straight line from a start state to a goal state.
   \item[2] \textit{Push}: The robot attempts to push an object $t$ from a start state to some goal state.
                           To compute such an action, we first use our generator to sample a robot pushing state.
                           Then an action that steers the robot to this state is concatenated with
                           an action provided by our policy for the same state.
                           Note that although this is a similar behavior as in our $\textsc{ExtendTree}$
                           function, the primitive does not propagate these actions through $\Gamma$.
                           Hence, any unintended contact that may occur during the approach motion is not
                           considered by this primitive.
\end{itemize}
For a fair comparison, we evaluate the algorithm for different choices of $k \in \{1, 4, 8\}$ and
$p_\mathit{rand}\in \{0, 0.25, 0.5, 0.75, 1\}$ and select the best.
Interestingly, the best performing choice for most test cases is $k=1$ and $p_\mathit{rand} = 0.0$.
This indicates that in our test cases there is not much being gained from performing random actions
rather than following the \textit{Push} primitive that is based on our policy and generator.

\subsubsection{PolicyGeneratorRRT}
To evaluate the effect of the slice-based exploration of our algorithm, we devise
a simplified version of our algorithm, PolicyGeneratorRRT. This algorithm
grows a tree on $\mathcal{C}_{0:m}$ similar to HybridActionRRT without
applying the concept of slices. In contrast to HybridActionRRT, however, it
applies the same $\textsc{ExtendTree}$ function as our algorithm.
Hence, the main difference to HybridActionRRT with the primitives defined above is that
an approach \textit{transit} and a following pushing action are propagated separately through $\Gamma$.
This allows the policy to be applied to the resulting state of the \textit{transit} action, i.e.\
it allows the policy, to some extent, to adapt to unintended contact during the approach motion.

\subsection{Technical Details}
We implemented the planning algorithms in C++ using OMPL \cite{sucan2012OMPL}.
Similarly, the simple generator and policy were also implemented in C++.
The learned generator and the learned policy in contrast were implemented in Python using PyTorch.
The communication between these and the planner was performed using Protobuf.
The overhead of this communication is not included
in our evaluation. As physics model we chose Box2D \cite{box2d}.
All experiments were run on Ubuntu 14.04 on an Intel(R) Core(TM) i7-3770 CPU@3.40GHz
with $32GB$ RAM\@.

\subsection{Evaluation of Policy and Generator}
\label{experiments:oracle}

\subsubsection{Quantitative Evaluation}
\begin{figure}[t]
        \setlength{\tabcolsep}{2pt}
        \begin{tabular}{cc}
                \begin{subfigure}[b]{0.24\textwidth}
                        \includegraphics[width=\textwidth]{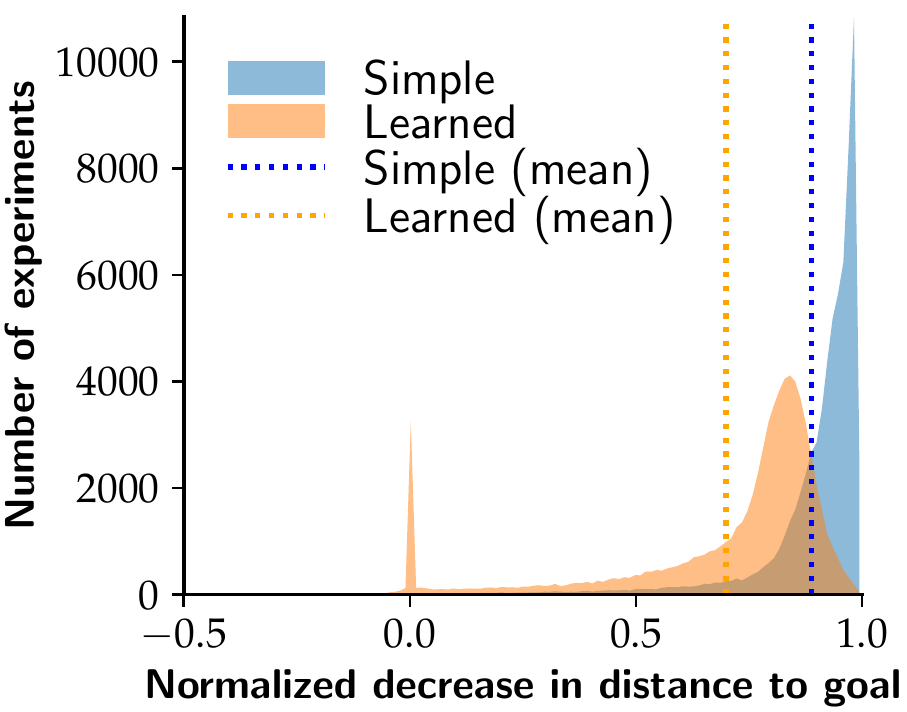}
                    \caption{Pushing results of a non-slippery object.}
                    \label{fig:1step-normal}
                \end{subfigure}
                &
                \begin{subfigure}[b]{0.24\textwidth}
                   \includegraphics[width=\textwidth]{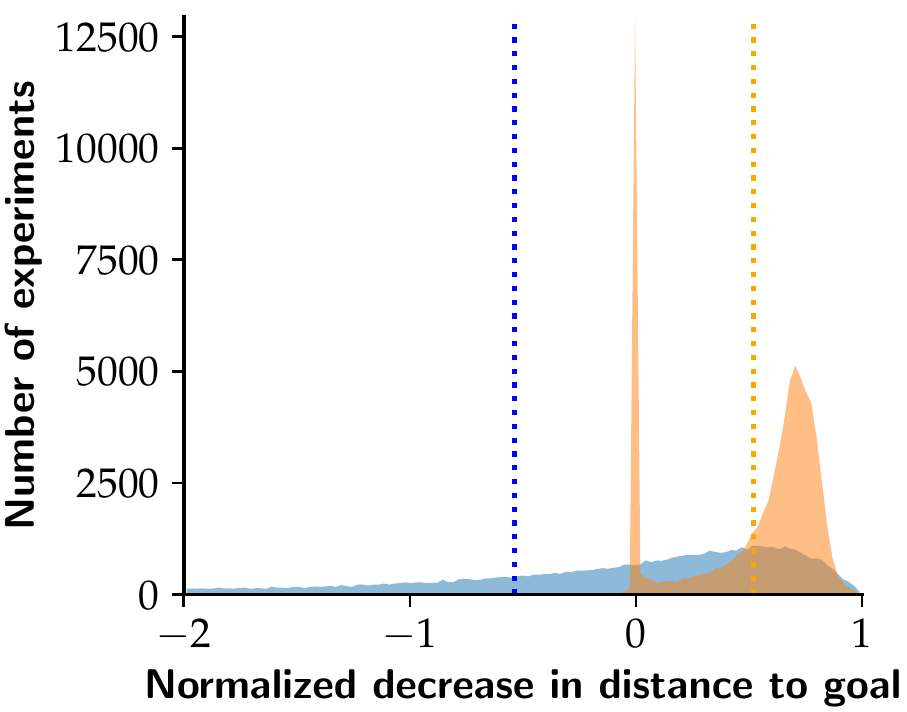}
                   \caption{Pushing results of a slippery object.}
                    \label{fig:1step-slippery}
                \end{subfigure}
        \end{tabular}
        \caption{
                Single push evaluations. We observe the Euclidean distance
                of an object to a random goal pose before and after executing
                the policy from a manipulation state. These histograms
                show the normalized decrease in distances, i.e.\ the
                distance after the execution divided by the distance before the execution.
                The best possible normalized decrease is $1.0$.
                \figref{fig:1step-normal} shows the result for an object that does not slide,
                whereas \figref{fig:1step-slippery} shows results for an object with lower mass
                and friction that does slide after being pushed.
        }
        \label{fig:oracle_evaluations}
\end{figure}

To evaluate our learned models, we compare their performance in transporting
a single object with that of the simple heuristic.
Our evaluation procedure follows the extension strategy of our planning
algorithm. Given an initial state of the object and a randomly sampled target state,
we first place the robot in a manipulation state produced by the generator.
Next, we query the policy to provide a single action to transport the object
towards the goal state and forward propagate this through the physics model
$\Gamma$. We then compare the Euclidean distance between the resulting object
state and the goal state.

\figref{fig:oracle_evaluations} shows the results of this procedure for two
different box-shaped objects that differ in mass and friction. One box is
slippery, i.e.\ it does not immediately come to rest after being pushed, and
one box is non-slippery.  In case of the non-slippery box, we observe that both
the learned and the simple models on average succeed at transporting the object
towards the target position. Although the learned models on average reduce the
distance by more than half, the simple models are yet better in this case. For
the slippery object, in contrast, the learned models are significantly better than
the simple ones. Here, the learned models achieve similar results as in the
non-slippery case, whereas the simple heuristic often results in a significant
increase in distance. This highlights a weakness of such simple hand-made heuristics.
A behavior that works well for some types of objects may not work well for others.
The learned models, on the other hand, are parameterized by the expected physical
properties of the object and can thus adjust their behavior. Achieving similar results
with hand-made heuristics would require significant engineering efforts.

\subsubsection{Qualitative Evaluation}
\begin{figure*}[t]
    \setlength{\tabcolsep}{2pt}
    \begin{subfigure}[b]{1.0\textwidth}
        \centering
        \includegraphics[width=1.0\textwidth]{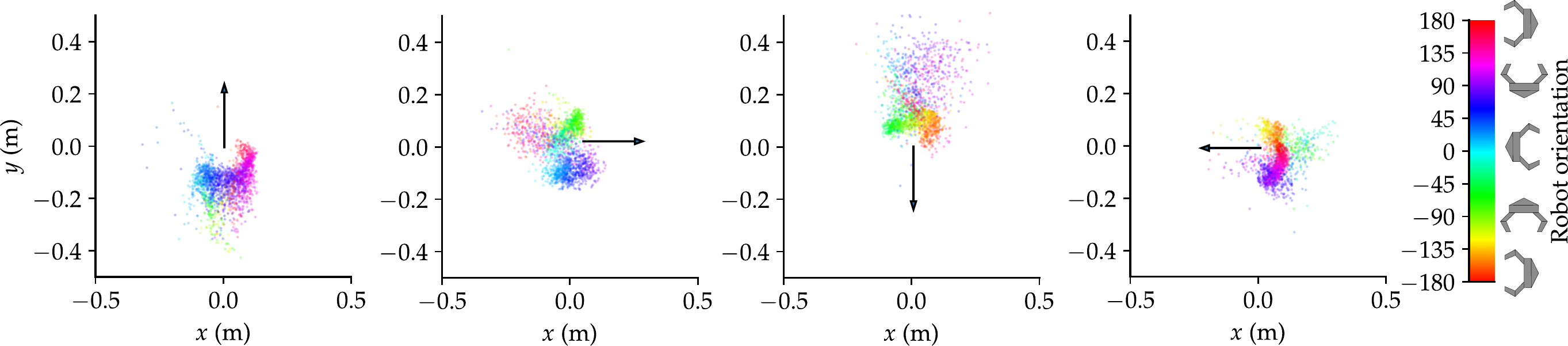}
        \caption{Learned manipulation states for a gripper-like robot. Most samples represent robot poses
        where the fingers are facing the object, and the robot is placed opposite to
        the pushing direction.}
        \label{fig:gan_samples:gripper}
    \end{subfigure}
    \begin{subfigure}[b]{1.0\textwidth}
        \centering
        \includegraphics[width=1.0\textwidth]{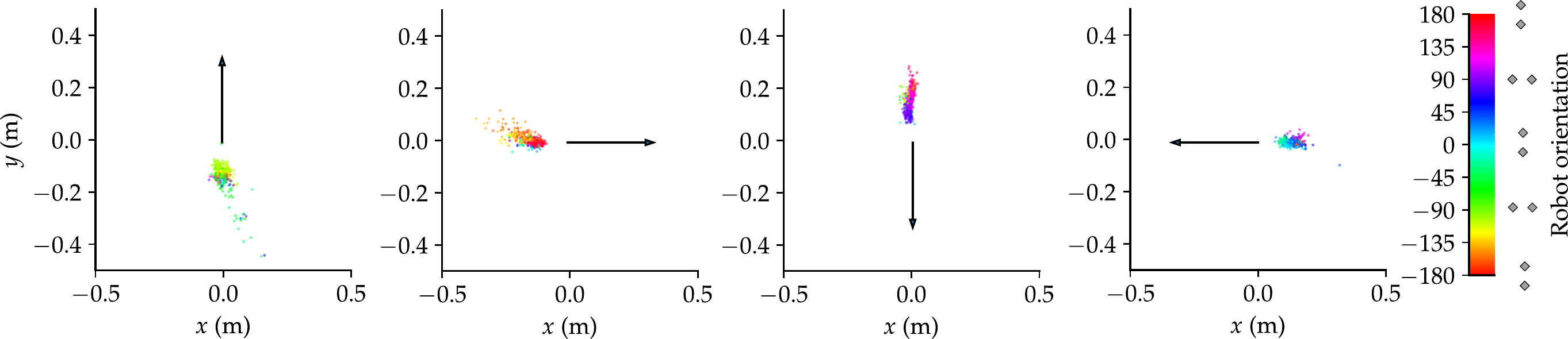}
        \caption{Learned manipulation states for a robot consisting of two squares.
            The learned behavior is to push such that the object is trapped between both squares.}
        \label{fig:gan_samples:two_point}
    \end{subfigure}
    \caption{Samples of robot states from our generative adversarial
    network. The arrow represents a desired
    push of an object starting in the origin. Each colored dot represents a robot position,
    where the color is the rotation of the robot. Consult our online supplementary material
    for a video showing how the samples are influenced by different arguments to the generator.
    }
    \label{fig:gan_samples}
\end{figure*}
Next, we qualitatively evaluate the learned generator.
\figref{fig:gan_samples} shows robot state samples produced by learned
generators for two different robots.  \figref{fig:gan_samples:gripper} shows
these samples for our gripper-shaped robot and
\figref{fig:gan_samples:two_point} shows these for a robot consisting
of two small squares. In both cases, the generators learned how the robot
should be positioned relative to the object in order to push it in some desired
direction. More interestingly, however, is that in both cases it also learned a
preference in orientation of the robot. In the case of the gripper, it prefers
orientations such that the object is placed between the gripper's fingers.  In
case of the two-point robot, it prefers orientations for which the robot
achieves two-point-contacts with the object. In both cases, these choices lead
to pushing behaviors that are more robust against uncertainty in object
properties than pushing from other orientations.

As can be seen, the samples show
that the generators learned distributions with a wide support over several possible
manipulation states. This is particularly useful, if some of these states are in collision
or can not be approached easily in the presence of obstacles. In particular for the gripper-shaped
robot, the generator learned states where the robot is facing perpendicular to the direction of transport.
Accordingly, the actions learned by the policy in these states, move the robot sideways.
Note that this is a behavior that makes specific use of the robot's geometric properties, i.e. its fingers.
One drawback of these wide distributions, however, is that there may
also be a small probability to sample distant states from which transporting the object
within one action is not possible. Such states are responsible for the cases
in \figref{fig:oracle_evaluations} where the distance to the goal is not decreased at all.

\begin{figure*}[ht]
        \includegraphics[width=0.48\textwidth]{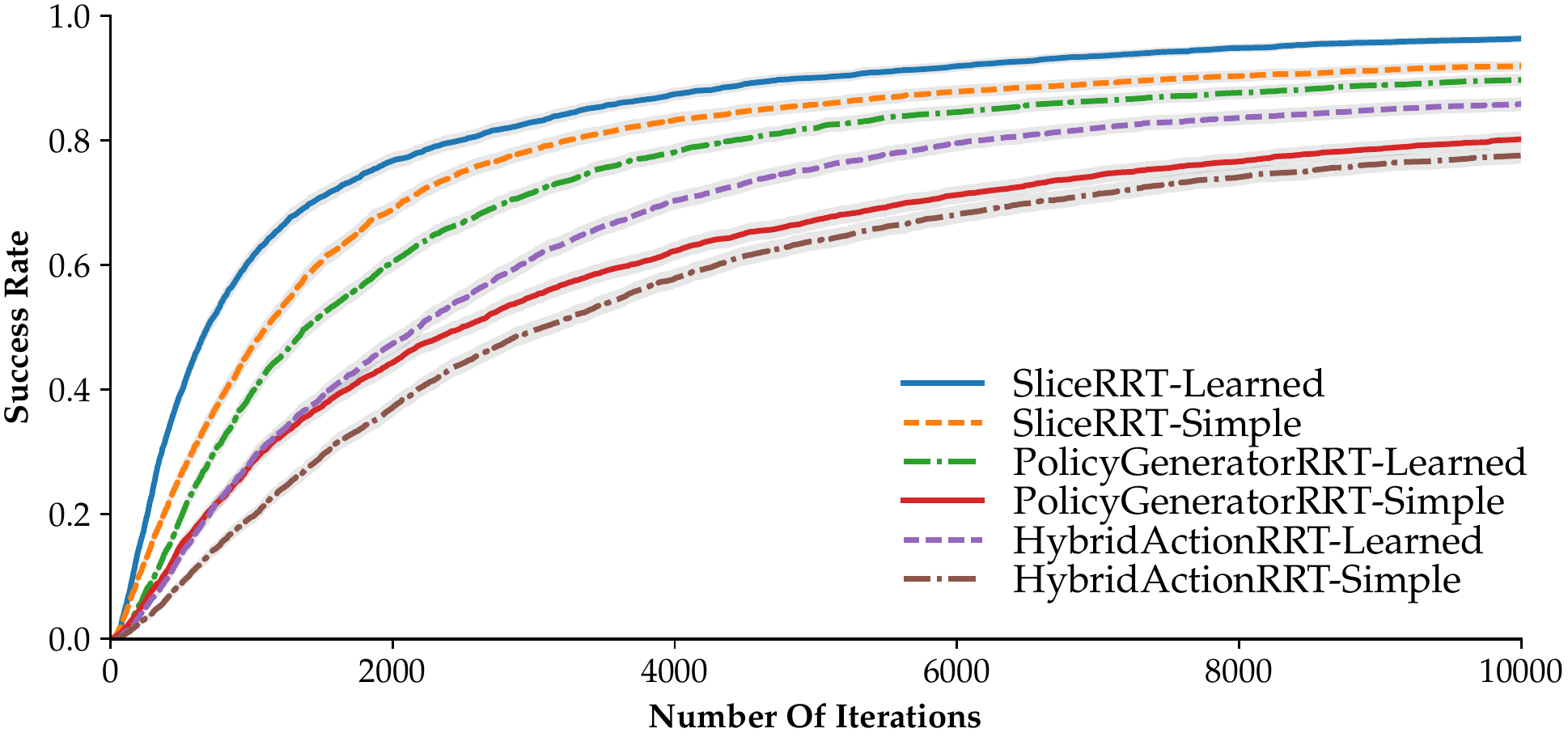}
        \includegraphics[width=0.48\textwidth]{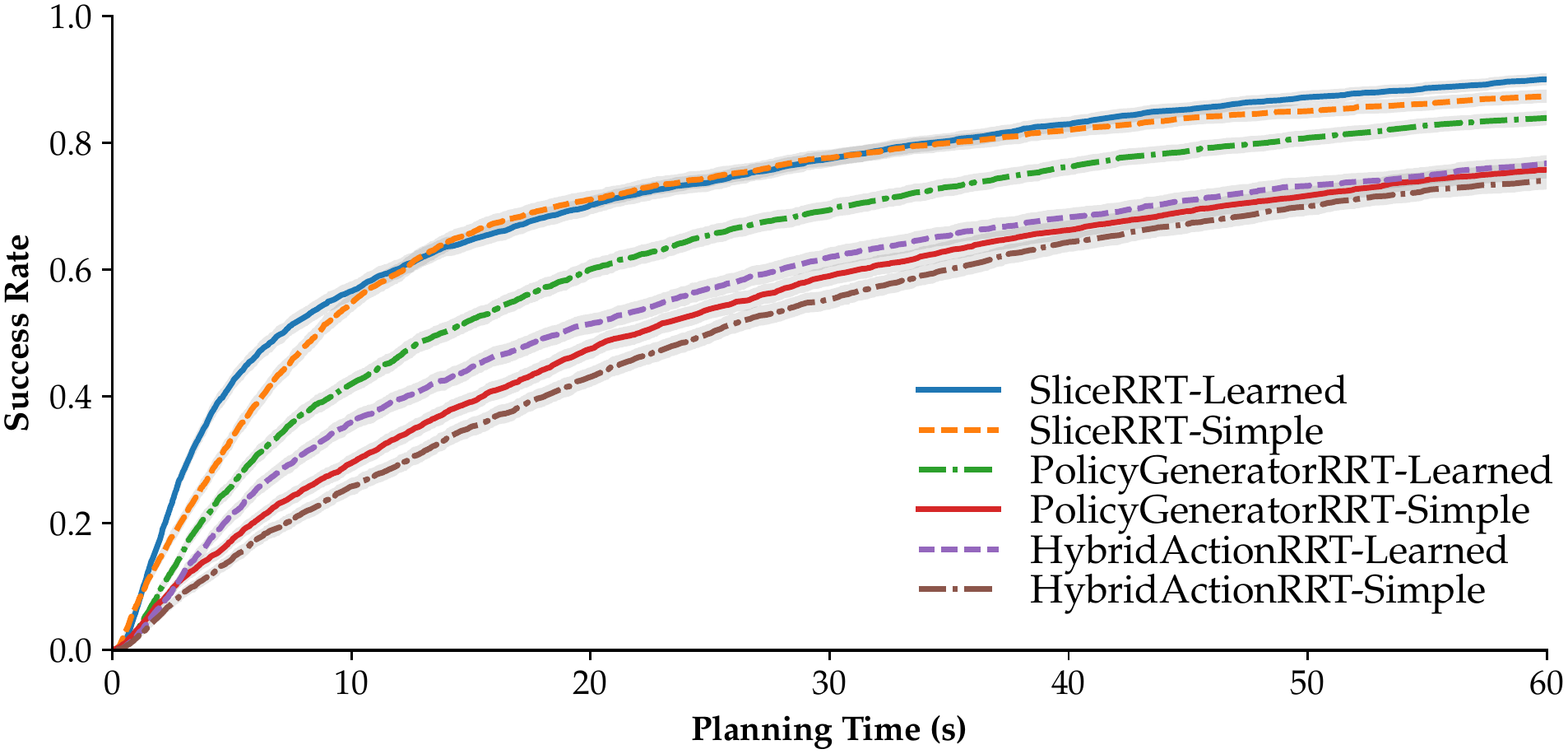}
        \caption{
                Planning success rate as function of number of iterations (\textit{left})
                and planning time (\textit{right}).
                For any given number of iterations $n$,
                the corresponding success rate can be interpreted as an empirically determined probability
                that an algorithm successfully finds a solution within $n$
                iterations. The shaded areas show the $95\%$ Wilson confidence
                interval of this probability. After a total of $180s$ the ratio of successful
                planning instances was:
                SliceRRT-Learned: $99.2\%$, SliceRRT-Simple: $96.6\%$,
                PolicyGeneratorRRT-Learned:$95.2\%$, PolicyGeneratorRRT-Simple: $91.9\%$,
                HybridActionRRT-Learned: $92.2\%$, HybridActionRRT-Simple: $91.2\%$.
                For SliceRRT and PolicyGeneratorRRT we chose $p_\mathit{rand} = 0.01$.
        }
        \label{fig:results:success_rates}
\end{figure*}

\subsection{Evaluation of the Planning Algorithm}
Next, we evaluate the effects of our different algorithm design choices as well as
how the learned models compare to the simple ones when used in a planner.
In the comparisons with our baselines, we refer to our algorithm by the name SliceRRT.

\subsubsection{Overall Performance}
We run all three algorithms with both the learned policy and generator as well
as the simple ones on six different scenes shown
in \figref{fig:scenes}. We run each algorithm-policy-generator combination for $640$ times on each scene
and record the runtime, the number of iterations and whether the algorithm succeeds at finding a solution
within a time budget of $180s$.

The planning success rates as a function of number of iterations are shown in
\figref{fig:results:success_rates} on the left.
As the number of iterations increases, more planning instances terminate successfully and the
success rates tend towards 1. Our SliceRRT algorithm achieves with both policy-generator models the steepest
initial increase as well as the highest success rates after the first \numprint{10000} iterations. The curves for
both PolicyGeneratorRRT and HybridActionRRT are significantly flatter for both the learned policy
and generator as well as the simple ones. Overall, the success rates remain below the one of SliceRRT.
It is notable that for each algorithm the learned policy and generator outperform the simple ones.

Similarly, the success rate as a function of planning time is
shown on the right of \figref{fig:results:success_rates} for the first $60s$.
Also here SliceRRT with learned policy and generator achieves the steepest initial increase and
highest final success rate. The difference, however,
between the simple and the learned policy and generator is smaller due to increased
computational costs of the learned models. We note, however, that this effect also depends on
implementation details.
Overall, these results indicate that:
\begin{itemize}
    \item[\textbf{1.}] The learned policy and generator provide better guidance than the simple ones;
    \item[\textbf{2.}] SliceRRT is more efficient in terms of iterations and runtime on the tested
            problems than the other algorithms.
\end{itemize}
In particular, the poorer performance of PolicyGeneratorRRT shows that
the slice-based exploration increases the algorithm's efficiency significantly.

\subsubsection{Performance per Scene}
\begin{figure}[t]
        \includegraphics[width=0.48\textwidth]{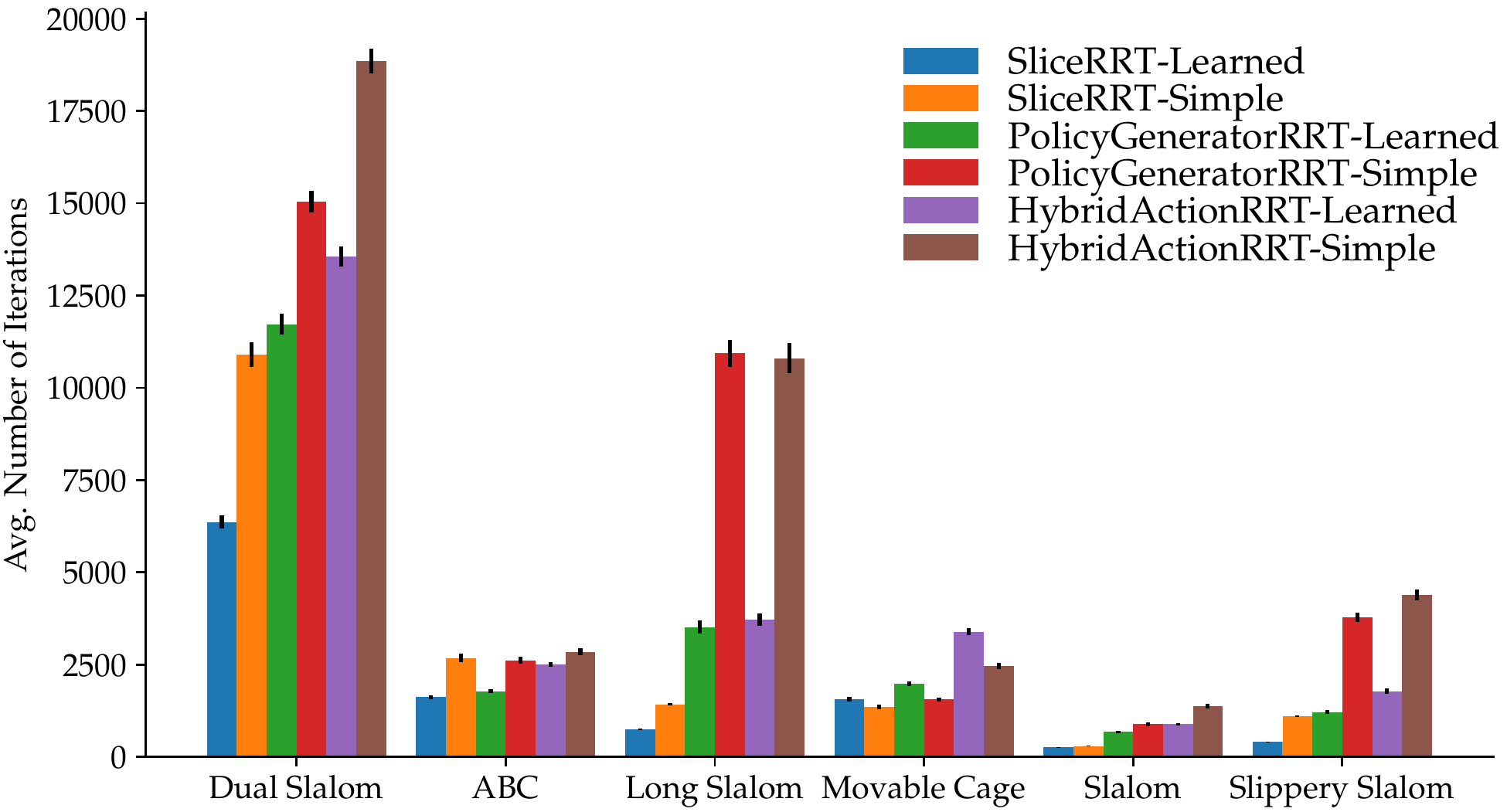}
        \includegraphics[width=0.48\textwidth]{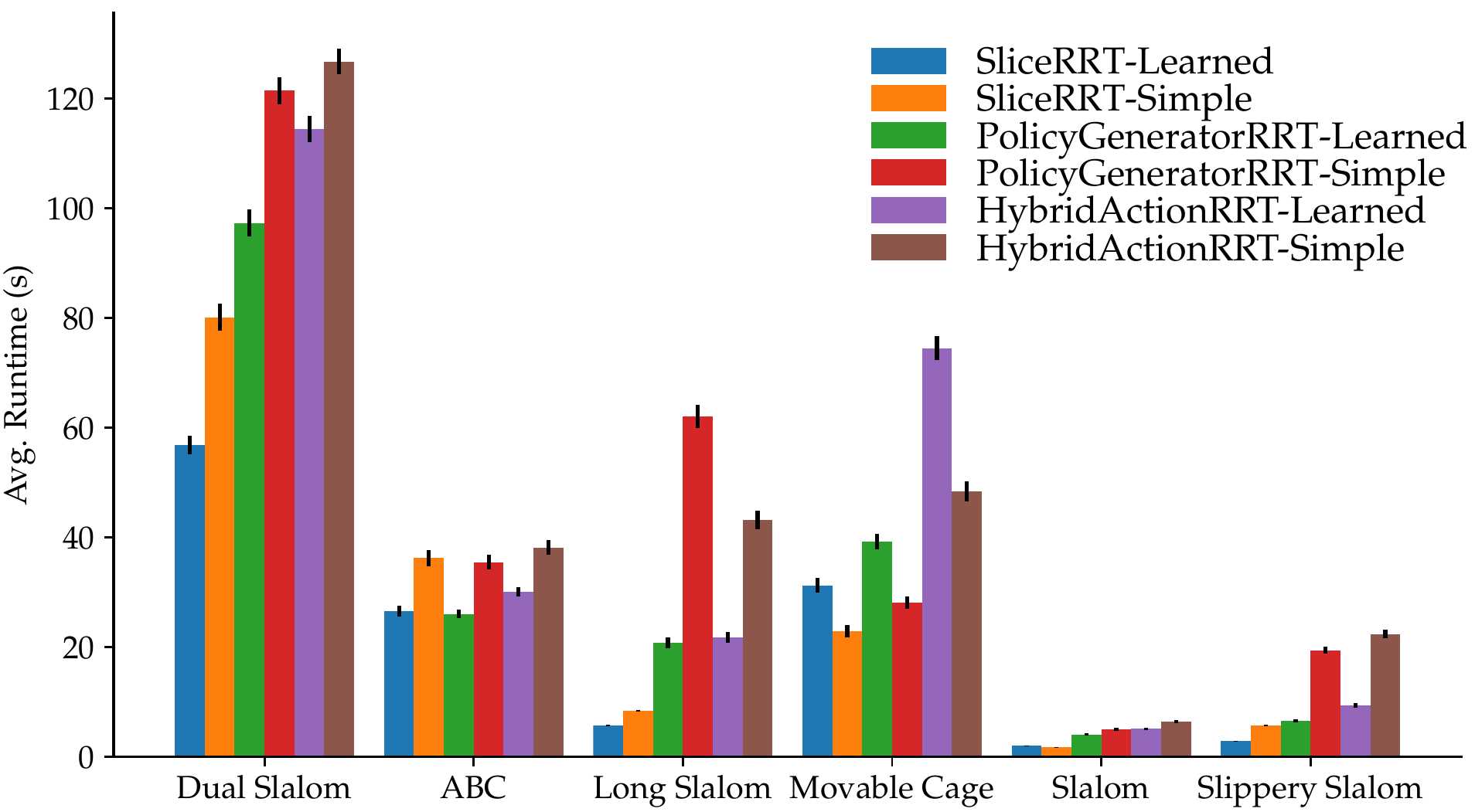}
        \caption{
                Average number of iterations (\textit{top}) and average runtime (\textit{bottom})
                per algorithm-policy-generator pair per tested scene.
        }
        \label{fig:results:per_env}
\end{figure}

Next, we investigate how the average runtime differs per test scene, see \figref{fig:results:per_env}.
For all algorithms we observe for most scenes better or similar results for the
learned policy and generator than for the simple ones. The differences between
the learned and the simple models are most significant on Dual Slalom,
ABC, Long Slalom and Slippery Slalom. This confirms our observations from \secref{experiments:oracle}
as all of these scenes contain at least one object with low
friction and low mass.

The best results on most scenes are achieved by SliceRRT with learned policy
and generator. The differences are particularly strong on the different slalom
scenes. In these scenes the robot and the objects can potentially be very
distant and separated by static obstacles. It appears that the slice-based
exploration has its strongest benefit in these situations. The differences
between PolicyGeneratorRRT and HybridActionRRT are not as significant for many
scenes. Most notably is the difference for Movable Cage, where we have many
movable objects, which are initially close to each other.  In such a situation
it is likely that the target object is moved through direct or indirect contact
when approaching a manipulation state. HybridActionRRT, in contrast to
PolicyGeneratorRRT, queries the policy before propagating the approach motion
to the manipulation state.  Choosing a pushing action given an updated state,
as in PolicyGeneratorRRT, seems in this case to be most beneficial.

    \section{Discussion And Conclusion}
\label{sec:discussion}
The goal of this work was to design an efficient algorithm
that can solve non-prehensile MAMO and rearrangement problems, while making
few limiting assumptions on the robot's manipulation abilities.
For this, we presented an algorithm based on the kinodynamic RRT algorithm
that explores the composite configuration space $\mathcal{C}_{0:m}$ of robot and objects.
The algorithm is agnostic to the robot's manipulation abilities
by modeling the effects of robot-centric actions through a dynamic physics model $\Gamma$.
It achieves efficiency by segmenting the search space into slices of similar object arrangements,
and deploying a learned pushing policy and robot state generator for guidance. The slice
segmentation allows the planner to select the most suitable states from its search tree
to extend towards a desired object arrangement. The learned robot state generator
provides the planner with robot states from which pushing an object in a desired direction
is possible. Similarly, the learned policy provides the planner with robot actions
achieving these pushes. Our experiments demonstrate that our approach can successfully
compute rearrangement solutions for various scenes without a human designer explicitly modeling the
robots manipulation abilities. Furthermore, all techniques together allow our algorithm
to explore its search space more efficiently than a comparable approach, leading to lower
average planning times.

\subsection{Slice-based Exploration}
The slice-based exploration structures the search and has several benefits over
a purely state-based exploration as applied in~\cite{King2015, Haustein2015, King2016}.
First, when selecting a state for extension, it naturally takes into account that the
planner operates on different subspaces of $\mathcal{C}_{0:m}$.
For many robots it is simpler to steer the robot between different states
within a slice than to compute actions that move objects between slices.
In a purely state-based approach, such as HybridActionRRT, this needs to be expressed
through a weighting factor that balances between robot and object state distances
when selecting a nearest neighbor.
In contrast, our slice-segmentation naturally reflects the difference between both types of states
and alleviates us from tuning such a weighting factor.

Second, the algorithm balances its effort on exploring \textit{transit}
actions within the different slices in a meaningful manner.
The subset $\mathcal{T} \cap s$ of explored states that lie
within the same slice $s$ form a sub-RRT for this slice\footnote{In principle, $\mathcal{T} \cap s$ may consist
of several disjoint connected components. This can occur if the robot first leaves the slice $s$ through
a \textit{transfer} action but later on enters the slice again through another \textit{transfer} action
that places the moved objects back to their original poses. We note, however, that this is
highly unlikely to occur.}.
This sub-RRT fulfills the task of classical robot path planning
given the object arrangement in $s$. It is extended every time $s$ is selected and
either expanded towards a uniformly sampled ($t=0$) or a manipulation state ($t \neq 0$).
The probability of this extension to occur is proportional to the measure of the Voronoi region
of $s$ w.r.t. $d_s$ in $\mathcal{S}$. Hence, the algorithm spends more time on exploring robot states
within slices for which not many neighboring object arrangements have been reached than for others.

\subsection{Forward Propagation through Physics Model}
As a physics-based approach, our approach shifts most assumptions on the manipulation abilities of the robot
and the mechanics of manipulation into the model $\Gamma$.
Treating this model as a black box has the benefit that it can be replaced with any physics model.
In our implementation we chose a rigid body physics simulator for this. Alternatively,
we could also choose, for instance, a learned model.
Assuming that $\Gamma$ is deterministic, however, implicitly assumes perfect
knowledge about the manipulated objects. In general, this is not feasible for
a robot operating in a real environment. As a consequence, solutions planned by our planner
have high chance of failure when executed on a real system due to inaccurate predictions
of $\Gamma$.

In the context of physics-based manipulation planning, this issue has recently been
addressed by several works~\cite{Johnson2016, Koval2015, King2017}. Koval et al.~\cite{Koval2015}
present a multi-arm-bandit-based meta-planner that selects the solution that is most robust
against model uncertainties from a set of solutions computed by a planner similar to ours.
King et al.~\cite{King2017} present a Monte-Carlo-Tree-Search-based algorithm that plans
robust non-prehensile MAMO solutions on belief space. In both works, our planner could be
applied as a primitive.

In future work, we intend to investigate how uncertainty in $\Gamma$ can be addressed further.
Here, we believe learning a pushing policy that is robust and provides uncertainty
reducing actions could be highly beneficial.

\subsection{Learning Pushing State Generator and Policy}
Learning the pushing state generator and policy from data generated using the physics model has
several advantages. First, retraining these allows an easy adaptation of the planner
to different robot embodiments and new object types. Second, we do not impose any unnecessary
restrictions on the robot's manipulation abilities. Third, both generator and policy
can be parameterized by expected physical properties of the objects and trained such
that the learned pushing behavior is robust against uncertainty in these quantities.

Training both policy and state generator for any robot, however, is challenging.
Our current approach to collecting training data has only been evaluated for robots with
configuration spaces in $SE(2)$. Due to the uniform random sampling, the procedure is
limited to robots with configuration and action spaces for which the probability of sampling
a robot state and action that pushes the target object is non-zero.
For robots with high degrees of freedom, e.g.\ a 7-DoF manipulator,
sampling such state-action pairs may either have zero or very low probability.
Hence, for such robots a more sophisticated data acquisition is required.
In future work, we plan to investigate this and extend our approach to robot manipulators.
Learning a pushing policy and state generator for such robots is particularly interesting as it
would allow the planner to purposefully utilize full-arm pushing actions that are difficult
to model otherwise.

We trained our pushing policy and generator to push a single object at a time.
While this doesn't restrict the planner to apply actions that push multiple objects at once,
learning a policy and a state generator for such multi-object pushing actions might be beneficial.
Another limitation of our current policy is its precision. While
it on average succeeds at reducing the distance to a desired state,
it does not succeed in reaching it exactly. In particular pushing an object towards
a desired orientation proved to be difficult for the trained policy.

Instead of training a one-step policy, we could alternatively integrate a
multi-step policy that operates as a closed-loop controller running in the physics model.
While this would likely improve the accuracy of pushing actions,
the additional computational overhead could prove disadvantageous.
In preliminary experiments, we equipped the algorithm with a local planner
for \textit{transit} actions that attempts to avoid collisions
when approaching a manipulation state. This version of our algorithm, however,
performed worse than the evaluated one that only applies straight line steering
for the robot. This indicates that there is a fundamental trade-off between
the versatility and computational overhead of local planners in the algorithm.

\subsection{Challenging Rearrangement Problems}
\begin{figure}[t]
    \center
    \includegraphics[width=0.33\textwidth]{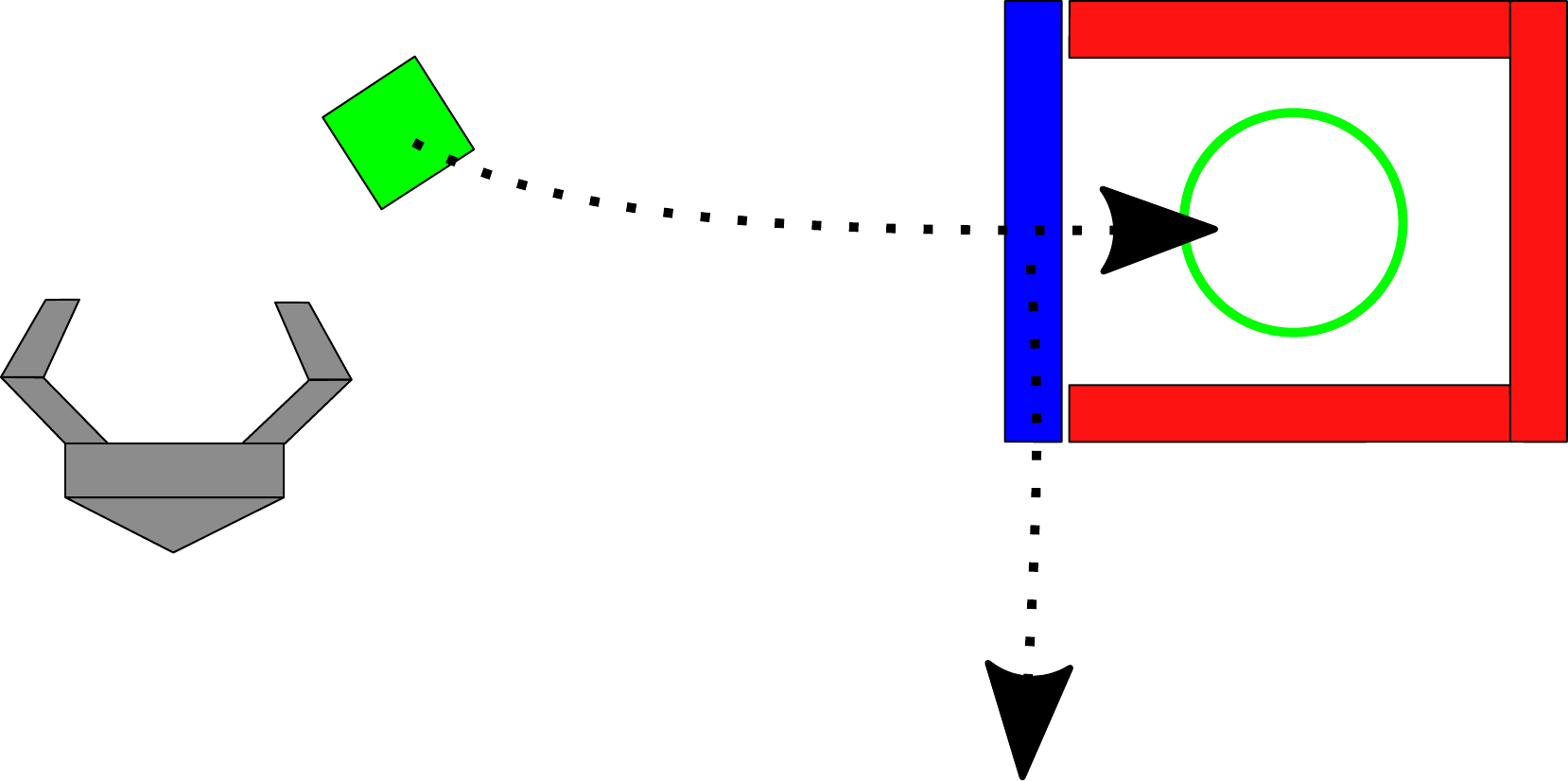}
    \caption{A problem with a narrow passage in object arrangement.
            To solve this problem the robot needs to push the blue object
            downwards and then push the green box into the goal region.}
    \label{fig:discussion:challenging_problem}
\end{figure}

A major challenge in rearrangement planning arises from the need to clear obstructions.
Consider the problem shown in \figref{fig:discussion:challenging_problem}, where
the robot first needs to push an obstacle aside to push its target object to the goal.
The directions in which this obstacle can be pushed are very limited due to the static obstacles (red).
In other words, a solution to this problem needs to pass through a narrow passage of
object arrangements, which has low probability to be sampled.

As a sampling-based approach, our planner struggles with scenarios like this.
A higher level logic is required that replaces the uniform slice
sampling with a more sophisticated mechanism. The challenge here lies in
formulating such a logic without making strong assumptions on the robot's manipulation abilities.
The question whether an object is obstructed by another depends not only on the
objects and their states, but also on the robot's embodiment.
Hence, we see a potential line of future work in learning a high level policy
that is conditioned on the robot's actual manipulation abilities
and directs the search to solve more challenging rearrangement problems.

\vspace{-0.05cm}

    \appendix
    \section{Appendix}
\subsection{Online Supplementary Material}
Supplementary material is available online on
\textbf{\url{https://joshuahaustein.github.io/learning_rearrangement_planning/}}.
The website contains videos of example solutions computed by
our planner, as well as a video illustrating the extension strategy of our algorithm
when using the generator and policy. Additionally, the website contains videos that show
how the learned policy and generator behave for different arguments, such as different
physical parameters of the target object and different target states.

\bibliographystyle{ieee/IEEEtran}
\bibliography{ieee/IEEEabrv,bibliography,rearrangement_works,pushing_works}
\end{document}